%% file: main.tex
\documentclass{article}

\usepackage{PRIMEarxiv}

\usepackage[utf8]{inputenc} 
\usepackage[T1]{fontenc}    
\usepackage{hyperref}       
\usepackage{url}            
\usepackage{booktabs}       
\usepackage{amsfonts}       
\usepackage{nicefrac}       
\usepackage{microtype}      
\usepackage{lipsum}
\usepackage{fancyhdr}       
\usepackage{graphicx}       
\graphicspath{{media/}}     

\usepackage{hyperref}
\usepackage{url}
\usepackage{booktabs}
\usepackage{amsmath}

\usepackage{amssymb}
\usepackage{xcolor}
\usepackage{graphicx}
\usepackage{tabularx}
\usepackage{wrapfig}
\usepackage{float}
\usepackage{multirow}
\usepackage{natbib}
\usepackage{titletoc}

\title{Invariant Atoms: Sparse Coordinates of Local Semantic Geometry in Language Model Representations}

\author{
Muhammad Ahtesham \quad
Xin Zhong
\\ [0.2em]
Department of Computer Science \\
University of Nebraska Omaha, Omaha, NE, USA \\[0.2em]
\texttt{\{mahtesham,xzhong\}@unomaha.edu}
}

\begin{document}
\maketitle

\vspace{-2.5em}
\begin{abstract}
\input{sections/Abstract}
\end{abstract}

\vspace{-1.0em}
\section{Introduction}
\label{sec:introduction}
\vspace{-0.75em}
\input{sections/Introduction}

\vspace{-1.5em}
\section{Related Work}
\vspace{-0.5em}
\label{sec:related_work}
\input{sections/Related}

\vspace{-1.5em}
\section{Invariant Atom Learning}
\label{sec:method}
\vspace{-1.0em}

\input{sections/Methodology}

\vspace{-1.50em}
\section{Experiments}
\label{sec:experiments}
\vspace{-1.00em}
\input{sections/Experiments}

\newpage
\bibliographystyle{unsrt}  
\bibliography{references}  

\newpage
\appendix
\section*{Appendix}
\input{sections/suplementery}

\end{document}

%% file: sections/Abstract.tex
Large language models often preserve meaning despite substantial changes in wording, style, and syntax, while small semantic edits can systematically alter their hidden representations. This suggests that semantic variation may be organized along recurring local directions. We propose the Invariant Atom Hypothesis: local semantic motion admits preferred sparse coordinates along directions that remain stable under meaning-preserving transformations. We learn a shared semantic frame and sparse coordinates that reconstruct semantic displacements while suppressing nuisance variation, with anchor-dependent diagonal modulation adjusting atom strengths without sample-specific rotations. Empirically, the atoms exhibit strong semantic--nuisance separation, sparse reconstruction, reproducible directions, and causal effects on model predictions. The learned geometry generalizes to unseen semantic neighborhoods and nuisance families, while local reweighting improves semantic selectivity and preserves a consistent global-to-local structure. Atom signatures also remain stable under model modification. These findings support reusable invariant directions as a sparse coordinate system for local semantic geometry in language models.


%% file: sections/Introduction.tex
\begin{wrapfigure}{r}{0.75\textwidth}
  \centering
  \vspace{-1.5em}
  \includegraphics[width=1.0\linewidth]{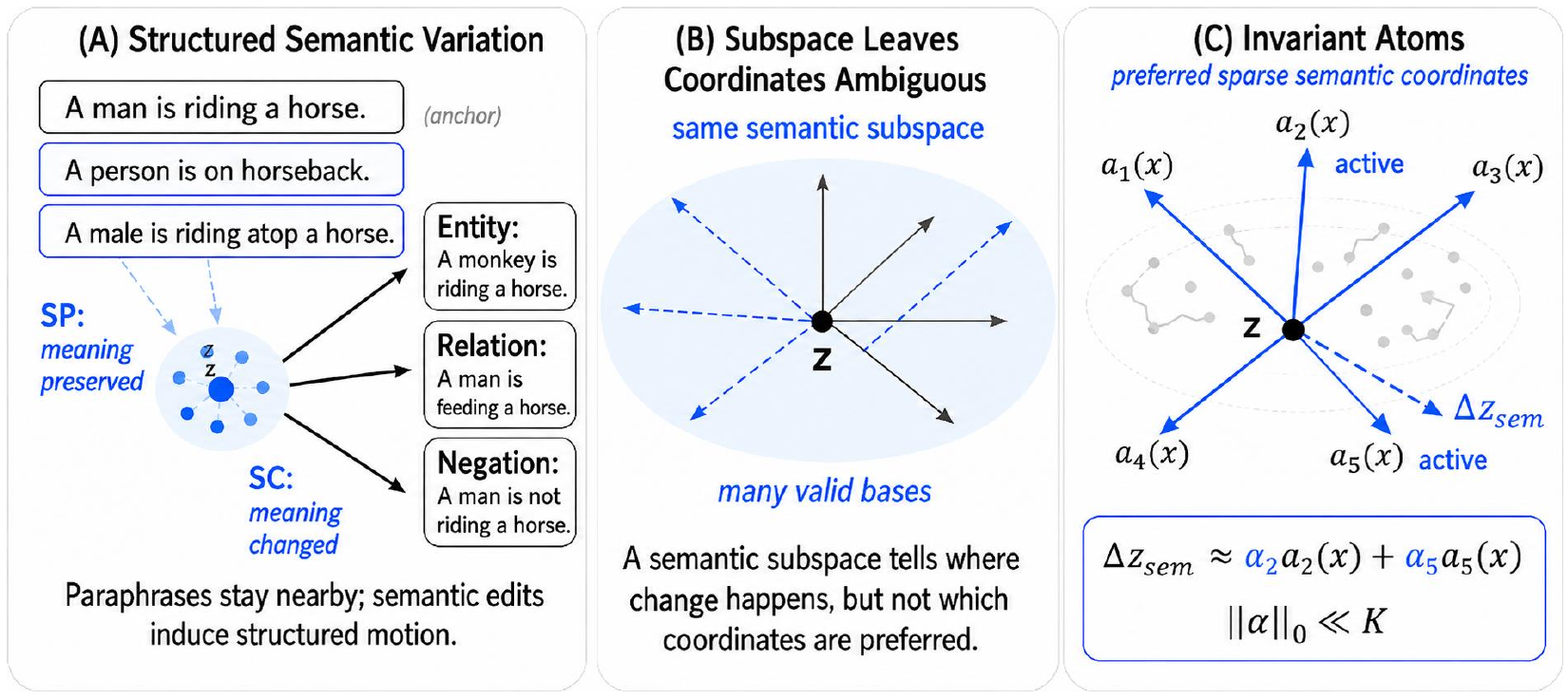}
  \vspace{-13.0em}
\caption{Invariant atoms as preferred coordinates of semantic variation.
(A) Meaning-preserving inputs remain locally stable, while semantic edits induce structured motion. (B) A semantic subspace identifies where such motion occurs but not a unique coordinate system. (C) Invariant atoms provide preferred sparse directions that compose semantic change while remaining stable to meaning-preserving variation.}
    \label{fig:FIA}
  \vspace{-1.00em}
\end{wrapfigure}

\vspace{-0.5em}
Large language models (LLMs) often preserve meaning despite substantial changes in wording, style, or syntax, while comparatively small changes in meaning can induce systematic shifts in their internal representations. This contrast raises a basic question about representation geometry: are semantic changes distributed arbitrarily throughout the hidden space, or are they organized along a recurring set of preferred directions? We study this possibility through invariant atoms, directional components that capture meaningful semantic motion while remaining stable to surface-level variation. As illustrated in Fig.~\ref{fig:FIA}, meaning-preserving variations remain within a local semantic neighborhood, whereas changes in meaning induce structured motion that may admit a sparse directional description.

\vspace{-0.5em}
Characterizing such directional structure requires more than identifying where semantic information is represented. High-dimensional hidden states simultaneously encode lexical form, syntax, style, context, and semantic content, so directions of large variation need not correspond to meaningful semantic factors. Moreover, even if semantic variation is concentrated within a lower-dimensional region, the region alone does not determine its coordinates: infinitely many rotated bases span the same subspace. The stronger question is therefore whether semantic motion itself has a preferred internal organization. If so, a useful representation should identify directions that respond consistently to changes in meaning, remain stable under meaning-preserving variation, and allow individual semantic changes to be expressed through only a small subset of those directions.

\vspace{-0.5em}
Existing approaches provide important pieces of this picture but do not jointly address this question. Sparse autoencoders and dictionary-learning methods decompose activations into sparse latent features, but primarily optimize reconstruction of the activation space rather than invariance defined by semantic transformations~\citep{bricken2023monosemanticity,huben2024sparse,templeton2024scaling}. Concept directions and activation-steering methods show that individual latent directions can encode or control meaningful properties~\citep{turner2023steering,li2023inference,lu2025sparse}, but typically identify directions associated with particular concepts or behaviors rather than a reusable coordinate system for semantic change. Local coordinate coding provides a mechanism for representing nonlinear manifolds using locally supported sparse coordinates~\citep{yu2009nonlinear}, but does not distinguish semantic motion from nuisance variation. These observations motivate a more selective question: does the local semantic geometry of language-model representations admit preferred sparse coordinates that remain stable under meaning-preserving transformations?

\vspace{-0.5em}
We answer this question by proposing and validating the Invariant Atom Hypothesis: meaningful semantic motion is organized by a preferred set of reusable directions, and individual semantic changes activate only a sparse subset of them. These atoms are invariant in the sense that meaning-preserving transformations may alter the representation locally while preserving the directional structure through which semantic change is expressed. 
Our contributions are threefold. 
(1) We formulate invariant atoms as preferred sparse coordinates of local semantic geometry, providing a geometric account of semantic variation that combines invariance, sparse composition, and approximately non-redundant directional structure. 
(2) We develop a learning framework that discovers these directions from meaning-preserving and meaning-changing variations. 
(3) We show that the resulting coordinates are reusable beyond the training perturbations: they generalize to held-out paraphrase styles and independently generated transformations, resolve diverse semantic change types, and support downstream analyses including transformation-robust semantic matching and stable representation signatures under model modification. 
Together, these results suggest that semantic variation in LLMs is organized not only within structured regions of representation space, but through reusable invariant directions that form a sparse local coordinate system.

%% file: sections/Related.tex
\vspace{-0.5em}
\noindent \textbf{Sparse representation learning and sparse autoencoders.}
Sparse representation learning seeks compact decompositions in which a small number of basis elements explain a high-dimensional signal, and recent sparse autoencoders (SAEs) extend this idea to language-model activations~\citep{bricken2023monosemanticity,huben2024sparse,templeton2024scaling,gao2025scaling}. In particular, $k$-sparse SAEs directly control activation sparsity through TopK selection while studying the tradeoff between reconstruction and feature quality~\citep{gao2025scaling}. Our formulation similarly combines learned directions with sparse coefficients, but targets semantic displacement and transformation-defined invariance. Rather than reconstructing an absolute activation $z$, invariant atoms reconstruct semantic displacement while explicitly suppressing semantic-preserving variation. Thus, the objective is not only sparse decomposition, but a transformation-defined separation between semantic and nuisance motion: directions selectively responsive to changes in meaning while remaining stable to surface-form variation.

\vspace{-0.5em}
\noindent \textbf{Semantic directions and dictionary representations in language models.}
A complementary line of work studies semantically meaningful directions in representation space through concept vectors, activation engineering, representation steering, and structured sparse representations~\citep{turner2023steering,li2023inference,lu2025sparse}. The linear representation hypothesis further formalizes concepts as directions associated with counterfactual changes and relates such directions to the geometry of representation space~\citep{park2024linear}, with subsequent work extending this view to categorical and hierarchical concept structure~\citep{park2025geometry}. Related approaches identify and remove attribute-associated subspaces~\citep{ravfogel2020null,belrose2023leace,ravfogel2022linear}. Collectively, these works support the view that semantic information can exhibit directional organization, but typically study directions associated with particular concepts, attributes, or behaviors. We instead ask whether semantic change itself admits a reusable coordinate system: directions that jointly span allowable semantic motion, compose individual changes sparsely, and remain invariant to meaning-preserving transformations. The resulting object is therefore a semantic frame rather than a set of independently discovered concept directions.

\vspace{-0.5em}
\noindent \textbf{Local coordinate coding and manifold representations.}
Local coordinate coding and related manifold methods model nonlinear data geometry through locally supported dictionary representations~\citep{yu2009nonlinear}. Their central principle is that a nonlinear manifold can be approximated locally and that nearby samples admit compact coordinates with respect to a suitable dictionary. Our formulation adopts the same geometric intuition but addresses a more selective problem. The atoms are not generic coordinates for reconstructing points on the representation manifold; they parameterize the directions of semantic motion around an anchor after semantic-preserving variation has been factored out. Consequently, invariant atoms combine three properties that are typically treated separately: local geometric structure, sparse coordinate composition, and invariance defined through semantic perturbations. They therefore parameterize an explicitly identified invariant semantic geometry rather than merely sparse latent features or generic local manifold coordinates.

%% file: sections/Methodology.tex
Section~\ref{sec:frame_geometry} introduces the geometric formulation of invariant atoms as local semantic frames. Section~\ref{sec:forward} defines the anchor-conditioned atom frame and sparse coordinate encoding, Section~\ref{sec:loss} introduces the geometric learning objectives, and Section~\ref{sec:training} presents the two-stage global-to-local optimization procedure.


\vspace{-1.25em}
\subsection{Invariant Atoms as Local Semantic Frames}
\label{sec:frame_geometry}
\vspace{-0.75em}

Let $\Phi_{\ell}$ denote a frozen language model at layer $\ell$, and let
\(
z=\Phi_{\ell}(x)\in\mathbb{R}^{d}
\)
denote the representation of an input $x$.
Following the manifold hypothesis~\citep{fefferman2016testing}, we assume that meaningful representations lie on a structured manifold
\(
\mathcal{M}\subset\mathbb{R}^{d}.
\)
Around an anchor $z\in\mathcal{M}$, the tangent space $T_z\mathcal{M}$ gives a first-order approximation to the directions along which the representation can locally vary while remaining consistent with the learned geometry. These directions need not share the same semantic role: some alter wording, style, or syntax while preserving meaning, whereas others correspond to changes in the represented semantic state. 
This distinction motivates a finer question than whether semantic information occupies a low-dimensional region: does the model organize local semantic change along a preferred set of directions? If semantic motion were distributed arbitrarily throughout $T_z\mathcal{M}$, its description would depend on an arbitrary choice of coordinates. We instead hypothesize that the local representation geometry contains recurrent directional structure, so that meaningful changes can be expressed through a small set of preferred semantic directions. 

\vspace{-0.5em}
\noindent \textbf{Semantic equivalence and invariant local geometry.}
Let $x\sim x'$ denote semantic equivalence, meaning that $x$ and $x'$ differ through a semantic-preserving transformation while expressing the same content. Locally, equivalent inputs trace directions that preserve semantic identity. We model these as a nuisance tangent component
\(
\mathcal{N}_z\subseteq T_z\mathcal{M},
\)
corresponding to motion within the same semantic equivalence class. Factoring out this motion leaves the local degrees of freedom associated with changes in meaning. 
Geometrically, this can be viewed as the local quotient
\(
T_z\mathcal{M}/\mathcal{N}_z.
\)
Under the ambient Euclidean metric and a local linear approximation, we represent this quotient by a complementary semantic component
\(
\mathcal{H}_z
\)
such that
\(
    T_z\mathcal{M}
    \approx
    \mathcal{N}_z \oplus \mathcal{H}_z .
\)
Here $\mathcal{H}_z$ captures the local degrees of freedom through which the model can change its encoded meaning after semantic-preserving variation has been removed. 
We seek an ordered collection of directions
\(
    A(x)
    =
    \big[
        a_1(x),\ldots,a_K(x)
    \big]
    \in\mathbb{R}^{d\times K},
\)
whose columns provide preferred coordinates for this semantic motion. An atom $a_k(x)$ is therefore an allowable local semantic direction: moving along it corresponds to a structured mode by which the model alters its internal representation of meaning around the anchor. Rather than relying on arbitrary ambient coordinates, the atoms posit a reproducible directional organization of local semantic geometry.

\begin{wrapfigure}{r}{0.7\textwidth}
  \centering
  \vspace{-1.5em}
  \includegraphics[width=1.0\linewidth]{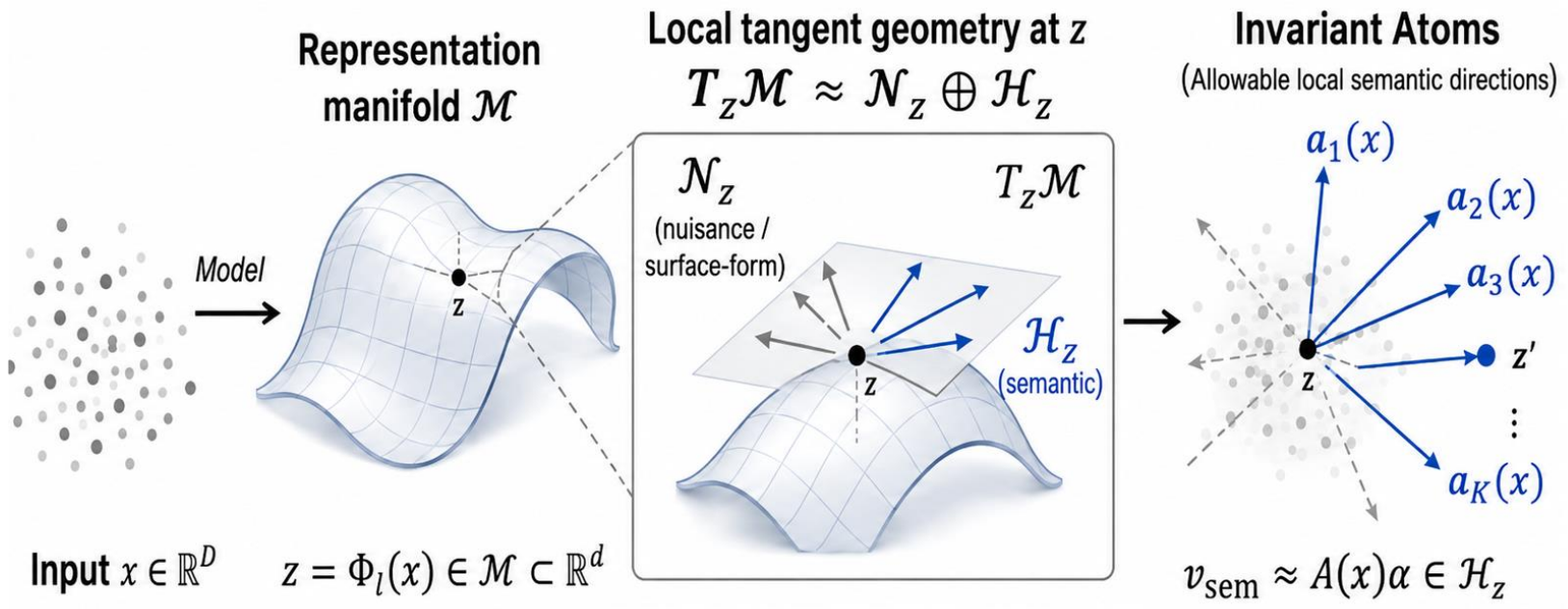}
  \vspace{-14.0em}
    \caption{Local geometric view of invariant atoms.
    Representations lie on a manifold $\mathcal{M}$ whose tangent space locally decomposes into nuisance directions $\mathcal{N}_z$ and semantic directions $\mathcal{H}_z$. Invariant atoms form preferred local directions within $\mathcal{H}_z$, allowing semantic motion to be expressed as a sparse combination of reusable coordinates.}
    \label{fig:theory}
  \vspace{-1.00em}
\end{wrapfigure}

\vspace{-0.5em}
The atom hypothesis further assumes that an individual semantic motion activates only
a small portion of this frame. For a local semantic tangent vector
$v_{\mathrm{sem}}\in\mathcal{H}_z$, we posit, for a sparsity budget $k\ll K$,
\(
    v_{\mathrm{sem}}
    \approx
    A(x)\alpha,
    \space
    \|\alpha\|_0 \le k \ll K .
\) 
Thus, the proposed structure is stronger than the existence of a semantic subspace.
A subspace specifies where semantic variation may occur; the atom frame further
specifies how such variation is composed, by organizing the local semantic geometry
into reusable directional elements and sparse coordinates. 
Fig.~\ref{fig:theory} summarizes the proposed idea. 
The term ``invariant" refers to the stability of these semantic directions under
semantic-preserving transformations. Although equivalent inputs may occupy slightly
different positions on $\mathcal{M}$, surface-level variation should not substantially
alter the semantic axes available around them. Let
\(
\bar{A}(x)
\)
denote the column-normalized frame associated with $A(x)$. For $x'\sim x$, we expect
\(
    d_{\mathrm{frame}}
    \big(
        \bar{A}(x),
        \bar{A}(x')
    \big)
    \approx 0 ,
\)
where $d_{\mathrm{frame}}$ measures discrepancy between ordered frames up to the
standard sign and permutation ambiguities of atoms. Thus, nuisance
transformations may change the representation itself while preserving the semantic
directional structure through which meaningful changes are expressed. The atoms are
therefore invariant not because their coordinates are context-independent, but because
their semantic interpretation remains stable under transformations that preserve meaning.

\vspace{-0.5em}
Our formulation is compatible with prior work on invariant semantic subspaces~\citep{dasgupta2026invariantfeatureslanguagemodels}, but targets a different geometric object. A local invariant subspace
\(
U_{\mathrm{inv}}(x)\subseteq T_z\mathcal{M}
\)
identifies a region of semantic variation that is relatively insensitive to semantic-preserving transformations, whereas invariant atoms introduce preferred coordinates within that region. Conceptually,
\(
\operatorname{span}\big(A(x)\big)\approx U_{\mathrm{inv}}(x),
\)
but the two objects are not equivalent: the subspace is unchanged under basis rotation, while the proposed atom hypothesis posits directions for which semantic changes admit compact, reusable, and sparse descriptions. Thus, invariant atoms refine invariant semantic structure from a subspace-level characterization into a semantic coordinate system.

\vspace{-0.5em}
\noindent \textbf{Near-Stiefel structure and non-redundant coordinates.}
A useful semantic frame should avoid degenerate or highly redundant directions. For the column-normalized frame
\(
\bar{A}(x)
=
[\bar{a}_1(x),\ldots,\bar{a}_K(x)],
\)
the ideal orthonormal case satisfies
\(
    \bar{A}(x)^\top\bar{A}(x)=I_K,
    \space
    \bar{A}(x)\in\mathrm{St}(K,d),
\)
where
\(
\mathrm{St}(K,d)
=
\{
A\in\mathbb{R}^{d\times K}:A^\top A=I_K
\}
\)
is the Stiefel manifold of ordered orthonormal $K$-frames. We do not assume that the representation manifold $\mathcal{M}$ is Stiefel, nor require exact orthogonality; rather, $\mathrm{St}(K,d)$ serves as a geometric reference for a well-conditioned frame. We therefore seek
\(
    \delta_{\mathrm{St}}(x)
    =
    \left\|
        \bar{A}(x)^\top\bar{A}(x)-I_K
    \right\|_F
    \ll 1 .
\) 
Sparse encoding alone does not prevent highly correlated atoms from representing similar directions. The mutual coherence $\mu(\bar A)=\max_{i\neq j}|\bar a_i^\top\bar a_j|$ measures this directional overlap. Lower coherence improves coefficient recovery and reduces ambiguity for a fixed atom frame~\citep{donoho2003optimally,tropp2004greed}. Thus, near-orthogonality complements sparse composition by encouraging distinct, well-conditioned coordinates. Reproducibility of the learned directions themselves is evaluated in Appendix~\ref{app:atom_reproducibility}.

\vspace{-0.5em}
Taken together, these considerations motivate the proposed \textbf{Invariant Atom Hypothesis}: the local semantic geometry of a language-model representation admits a preferred, approximately orthogonal frame that is stable under semantic-preserving variation, and semantic motion can be expressed through sparse combinations of these directions. Invariant atoms therefore provide more than a low-dimensional semantic subspace; they define local coordinates for systematic changes in the model's internal representation of meaning. The following sections develop a data-driven procedure for learning these atom frames.



\vspace{-1.25em}
\subsection{Atom Composition and Sparse Encoding}
\label{sec:forward}
\vspace{-0.75em}

\begin{wrapfigure}{r}{0.75\textwidth}
  \centering
  \vspace{-1.5em}
  \includegraphics[width=1.0\linewidth]{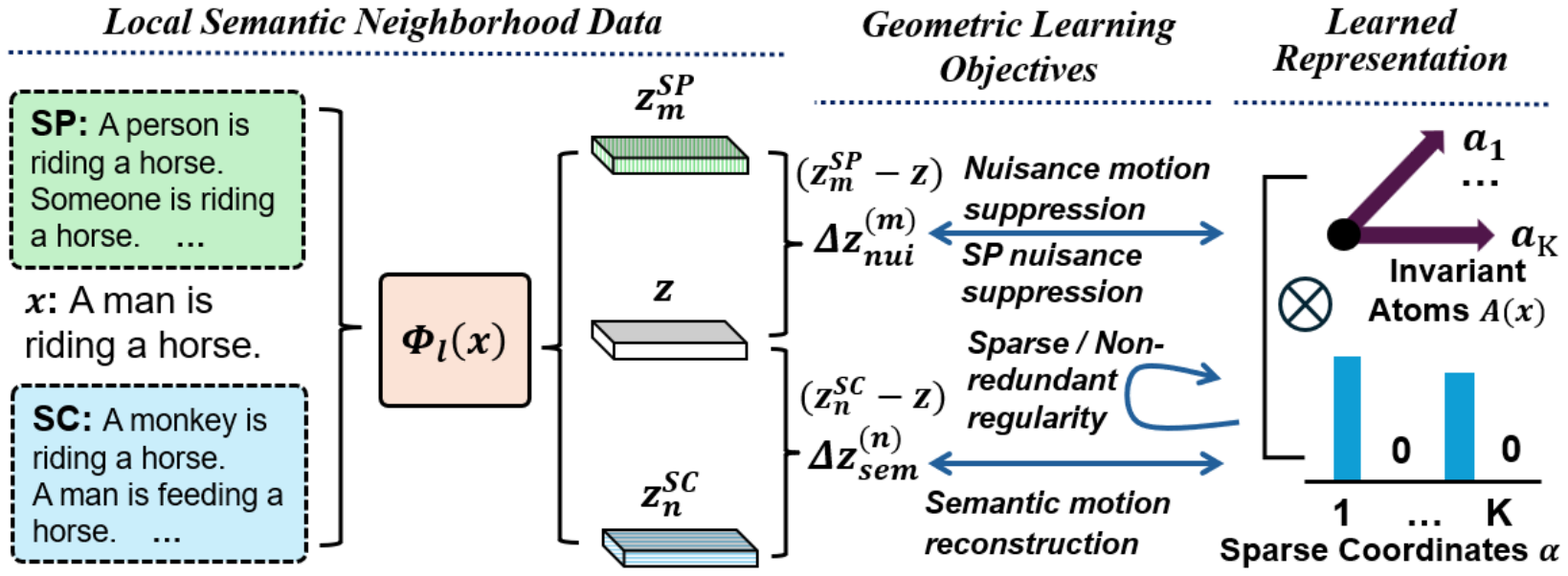}
  \vspace{-15.0em}
    \caption{Learning invariant atoms from local semantic variation.
    SP and SC transformations define nuisance and semantic displacements around an anchor representation. Geometric objectives suppress nuisance motion, preserve SP consistency, and reconstruct semantic motion, yielding sparse and non-redundant atoms.}
    \label{fig:Method}
  \vspace{-1.0em}
\end{wrapfigure}

We instantiate the Invariant Atom Hypothesis by learning a semantic frame around each anchor representation and sparse coordinates for local semantic motion. As summarized in Fig.~\ref{fig:Method}, semantic-preserving (SP) and semantic-changing (SC) transformations provide two complementary probes of the local representation geometry: SP variations characterize nuisance motion that should be suppressed by the learned frame, whereas SC variations provide semantic motion that the atoms should reconstruct. 

\vspace{-0.5em}
\noindent \textbf{Local representation variations.}
Let $x$ denote an anchor input with representation $z=\Phi_{\ell}(x)\in\mathbb{R}^{d}$. 
In our experiments, $\Phi_\ell$ includes fixed per-dimension feature
standardization using corpus-level statistics that do not depend on the
SP/SC labels (see Appendix~\ref{app:implementation}). 
For its $m$-th SP transformation $x_m^{\mathrm{SP}}$ and $n$-th SC transformation $x_n^{\mathrm{SC}}$, we denote their representations by
\(
z_m^{\mathrm{SP}}=\Phi_{\ell}(x_m^{\mathrm{SP}})
\)
and
\(
z_n^{\mathrm{SC}}=\Phi_{\ell}(x_n^{\mathrm{SC}}).
\)
Their finite representation displacements relative to the anchor are
\(
\Delta z_{\mathrm{nui}}^{(m)}
=
z_m^{\mathrm{SP}}-z
\)
and
\(
\Delta z_{\mathrm{sem}}^{(n)}
=
z_n^{\mathrm{SC}}-z.
\)
These displacements provide empirical probes of the local geometry introduced in Section~\ref{sec:frame_geometry}: $\Delta z_{\mathrm{nui}}^{(m)}$ primarily reflects variation that preserves semantic identity, whereas $\Delta z_{\mathrm{sem}}^{(n)}$ reflects motion induced by a change in meaning. We use the generic notation $\Delta z\in\mathbb{R}^{d}$ when the distinction is not required.

\vspace{-0.5em}
\noindent \textbf{Anchor-conditioned atom frame.}
Let
\(
A_0=[a_1,\ldots,a_K]\in\mathbb{R}^{d\times K}
\)
denote a shared base semantic frame. As detailed in Section~\ref{sec:training}, $A_0$ is first learned across training examples and subsequently used to initialize an anchor-conditioned frame. Given the anchor representation $z$, a shared modulation network $g_{\eta}$ predicts atom-wise adjustments
\(
r(x)
=
\tanh\!\left(g_{\eta}(z)\right)
\in\mathbb{R}^{K}.
\)
Thus, $g_{\eta}$ operates directly on the latent anchor $z$, while $r(x)$ denotes the resulting input-conditioned modulation. We define
\(
D(x)
=
\operatorname{diag}\!\left(1+r(x)\right)
\)
and construct the anchor-conditioned semantic frame as
\(
A(x)
=
A_0D(x).
\)
Equivalently, the $j$-th atom is
\(
a_j(x)
=
\left(1+r_j(x)\right)a_j,
\ j=1,\ldots,K.
\)
The shared frame $A_0$ therefore determines the reusable semantic directions, while $D(x)$ modulates their relative strengths according to the local semantic context. Because the modulation is diagonal, $A(x)$ preserves the directions of the shared atoms rather than introducing arbitrary input-dependent rotations.

\vspace{-0.5em}
\noindent \textbf{Sparse coordinate encoding.}
For a local displacement $\Delta z\in\mathbb{R}^{d}$, we first compute an unconstrained coordinate vector through a learned encoder,
\(
\widetilde{\alpha}(\Delta z)
=
W\Delta z+b,
\quad
W\in\mathbb{R}^{K\times d},
\quad
b\in\mathbb{R}^{K}.
\)
We then retain the $k$ entries with largest absolute magnitude,
\(
\alpha(\Delta z)
=
\operatorname{TopK}
\left(
\widetilde{\alpha}(\Delta z),k
\right),
\)
with all remaining entries set to zero. Hence,
\(
\|\alpha(\Delta z)\|_0\leq k\ll K,
\)
so each local displacement is represented using only a small subset of the available atoms. The retained coefficients remain signed, allowing an atom direction to contribute in either orientation. The encoder parameters $W$ and $b$ are optimized separately from the atom frame; as described in Section~\ref{sec:training}, $W$ is initialized from $A_0^\top$ and subsequently learned jointly with the remaining model components.

\vspace{-0.5em}
\noindent \textbf{Sparse semantic composition.}
For an SC displacement $\Delta z_{\mathrm{sem}}^{(n)}$, the selected coordinates define its atom-based reconstruction
\(
\widehat{\Delta z}_{\mathrm{sem}}^{(n)}
=
A(x)\alpha\!\left(\Delta z_{\mathrm{sem}}^{(n)}\right)
=
\sum_{j\in\mathcal{I}_k^{(n)}}
\alpha_j^{(n)} a_j(x),
\)
where $\mathcal{I}_k^{(n)}$ denotes the active Top-$k$ index set and
\(
\alpha^{(n)}
=
\alpha(\Delta z_{\mathrm{sem}}^{(n)}).
\)
This decomposition separates two aspects of the local semantic geometry: $A(x)$ specifies the semantic directions available around the anchor, while $\alpha^{(n)}$ specifies which of those directions, and with what signed strengths, compose a particular semantic change. 
The same frame $A(x)$ is evaluated on SP displacements $\Delta z_{\mathrm{nui}}^{(m)}$, but these variations are not reconstructed as semantic motion. Instead, their responses along the learned atom directions are suppressed, while semantically equivalent inputs are encouraged to maintain consistent atom-space coordinates. Together with sparse and structural regularization, these constraints define the geometric learning objectives introduced next.


\vspace{-1.25em}
\subsection{Geometric Learning Objectives}
\label{sec:loss}
\vspace{-1.0em}

The forward model defines an anchor-conditioned frame $A(x)$ and sparse coordinates $\alpha$, but not which directions should become invariant atoms. We therefore learn the frame using complementary objectives that reconstruct SC motion, suppress SP motion, preserve coordinate consistency, and discourage redundant or collapsed atoms.

\vspace{-0.5em}
\noindent \textbf{Semantic reconstruction.}
For the $n$-th semantic-changing displacement $\Delta z_{\mathrm{sem}}^{(n)}$, let
\(
\alpha_n
=
\alpha\!\left(\Delta z_{\mathrm{sem}}^{(n)}\right)
\)
denote its sparse coordinates. We require the selected atoms to reconstruct the corresponding semantic motion:
\(
    \mathcal{L}_{\mathrm{sem}}
    =
    \frac{1}{N}
    \sum_{n=1}^{N}
    \left\|
        \Delta z_{\mathrm{sem}}^{(n)}
        -
        A(x)\alpha_n
    \right\|_2^2 ,
\)
where $N$ is the number of SC transformations associated with the anchor. This objective aligns the learned atoms with directions that explain changes in meaning rather than simply directions of large representation variance.

\vspace{-0.5em}
\noindent \textbf{Nuisance motion suppression.}
For the $m$-th semantic-preserving displacement $\Delta z_{\mathrm{nui}}^{(m)}$, invariance requires that surface-form variation induce little response along the semantic frame. We therefore minimize
\(
    \mathcal{L}_{\mathrm{nui}}
    =
    \frac{1}{M}
    \sum_{m=1}^{M}
    \left\|
        A(x)^\top
        \Delta z_{\mathrm{nui}}^{(m)}
    \right\|_2^2 ,
\)
where $M$ is the number of SP transformations associated with the anchor. Together, $\mathcal{L}_{\mathrm{sem}}$ and $\mathcal{L}_{\mathrm{nui}}$ operationalize the central selectivity of invariant atoms: the learned directions should explain semantic-changing motion while remaining insensitive to semantic-preserving variation.

\vspace{-0.5em}
\noindent\textbf{SP nuisance suppression.}
The implementation includes a second SP penalty,
\(
\mathcal{L}_{\mathrm{stress}}
=
\frac{1}{M}\sum_{m=1}^{M}
\left\|A(x)^\top z-A(x)^\top z^{\mathrm{SP}}_m\right\|_2^2.
\)
Since $z^{\mathrm{SP}}_m-z=\Delta z_{\mathrm{nui}}^{(m)}$, we have
$\mathcal{L}_{\mathrm{stress}}=\mathcal{L}_{\mathrm{nui}}$ for both Global and Local. The two weighted terms therefore jointly control the strength of SP suppression.

\vspace{-0.5em}
\textbf{Sparse and non-redundant frame regularity.} 
The Top-$k$ encoder limits each displacement to at most $k$ active atoms.
To discourage redundant directions, we additionally use
\(
  \mathcal{L}_{\mathrm{orth}}
  = \big\| A(x)^{\top} A(x) - I_K \big\|_F^2 ,
\)
applied to the frame used for each anchor, so that $A(x) = A_0$ for the
Global model. For the Local model, the penalty also keeps the modulated
atom norms close to one; because the modulation is diagonal, the atom
directions remain those of the shared frame $A_0$.

\vspace{-0.5em}
\textbf{Combined objective.}
The learning objective is
\(
\mathcal{L}
=
\lambda_{\mathrm{sem}}\mathcal{L}_{\mathrm{sem}}
+
\lambda_{\mathrm{nui}}\mathcal{L}_{\mathrm{nui}}
+
\lambda_{\mathrm{stress}}\mathcal{L}_{\mathrm{stress}}
+
\lambda_{\mathrm{orth}}\mathcal{L}_{\mathrm{orth}}.
\)
Both training stages use these objectives, with a shared
frame in Stage 1 and an anchor-conditioned frame in Stage 2.
Fixed coordinate-normalization factors are absorbed into
the weights; the implemented reductions and weights
are specified in Appendix~\ref{app:atom_training}.



\vspace{-1.25em}
\subsection{Two-Stage Global-to-Local Optimization}
\label{sec:training}
\vspace{-1.0em}

The semantic frame must capture structure at two scales: directions that recur across the dataset and their context-dependent relevance around a particular anchor. We therefore use a two-stage optimization procedure. Stage~1 learns a shared semantic frame, while Stage~2 initializes from this solution and jointly refines the shared frame together with an anchor-dependent modulation. Both stages use the objective in Section~\ref{sec:loss}.

\vspace{-0.5em}
\noindent \textbf{Stage 1: learning a global semantic frame.}
We first learn a single frame
\(
A_0=[a_1,\ldots,a_K]\in\mathbb{R}^{d\times K}
\)
shared across all anchors, with
\(
A(x)=A_0.
\)
The matrix $A_0$ is a directly trainable parameter and is updated by backpropagation jointly with the sparse encoder parameters $W$ and $b$. Thus, the same atom directions must reconstruct semantic displacements while suppressing nuisance variation across heterogeneous local neighborhoods. 
We initialize $A_0$ with the leading $K$ principal
directions of the training SC displacements, which form an orthonormal
frame, and set $W = A_0^{\top}$ and $b = 0$.
After initialization, $A_0$ and $W$ are optimized independently. Stage~1 therefore learns a reusable global semantic frame rather than keeping the atom frame fixed at its initialization.

\vspace{-0.5em}
\noindent \textbf{Stage 2: adapting the frame to local semantic context.}
We initialize Stage~2 from the learned Stage~1 parameters and introduce the modulation network $g_\eta$. Given
\(
z=\Phi_\ell(x),
\)
we compute
\(
r(x)=\tanh(g_\eta(z))
\),
\(
D(x)=\operatorname{diag}(1+r(x))
\),
and
\(
A(x)=A_0D(x).
\)
The final layer of $g_\eta$ is zero-initialized, so Stage~2 begins with
\(
A(x)=A_0
\)
and gradually introduces anchor-dependent modulation. 
During Stage~2, $A_0$, $W$, $b$, and $g_\eta$ are optimized jointly. Hence, the shared frame is allowed to co-adapt globally with the modulation network rather than remaining fixed at its Stage~1 solution. 
Importantly, however, the input-dependent adaptation remains diagonal: different anchors may reweight the learned directions, but cannot introduce independent sample-specific rotations. The resulting factorization separates a globally shared directional structure, represented by $A_0$, from its locally varying expression through $D(x)$.

\vspace{-0.5em}
The semantic frame should capture reusable directions while allowing their local relevance to vary with context. To prevent the local modulator from absorbing anchor-specific variation before a coherent shared frame emerges, Stage~1 first learns a global coordinate system shared across anchors, and Stage~2 then refines it through anchor-specific diagonal scaling. This continuation biases the model toward globally reusable directions with locally adaptive relevance.

%% file: sections/Experiments.tex
We evaluate whether invariant atoms emerge consistently across model families, causally influence predictions, and generalize beyond the transformations and semantic neighborhoods.


\vspace{-1.25em}
\subsection{Data and Models}
\label{sec:data_models}
\vspace{-0.25em}

\vspace{-0.5em}
\noindent \textbf{SP/SC neighborhood construction.}
Our method is learned from controlled SP and SC transformations around anchor inputs. The primary dataset contains $1{,}400$ anchor-centered semantic neighborhoods and $64{,}093$ sentences in total, covering $274$ topics across $13$ broad domains, including animals, science, countries, technology, food, history, mathematics, arts, health, philosophy, sports, and environment. Each neighborhood contains one anchor together with approximately $24$ SP and $21$ SC variants on average. SP variants preserve propositional content while varying lexical form, syntax, and style across eight transformation families: formal, casual, question, technical, simplified, verbose, passive, and imperative. SC variants modify meaning through seven controlled transformation types: entity, attribute, relation, negation, quantity, temporal, and intent. Thus, SP and SC samples provide empirical nuisance and semantic displacement vectors, respectively, corresponding to $\Delta z_{\mathrm{nui}}$ and $\Delta z_{\mathrm{sem}}$ in Section~3. Figure~1 illustrates this distinction: surface-form changes preserve the semantic state, whereas controlled entity, relation, or negation edits induce localized semantic motion. 
For the standard experiments, even-indexed SP/SC variants are used for training and odd-indexed variants for evaluation, with no overlap between the two sets; generalization beyond known anchor neighborhoods is evaluated separately in Section~\ref{sec:generalization}.

\vspace{-0.5em}
\noindent \textbf{Models analyzed.}
We evaluate seven model families spanning different architectures and training paradigms: Mistral-7B-Instruct-v0.3, LLaMA-3.1-8B-Instruct, Gemma-2-9B, Qwen2.5-7B-Instruct, GLM-4-9B-Chat, DeepSeek-MoE-16B-Chat, and Falcon3-7B. This diversity tests whether invariant atom structure transfers across model families rather than depending on a particular architecture or training recipe. For each model, we extract final-token hidden states at the target layer and learn the atom frame independently in that representation space. Unless otherwise specified, we use 256 atoms with top-32 sparsity; layer-wise and sparsity analyses are reported separately.



\vspace{-1.25em}
\subsection{Empirical Evidence for Invariant Atoms}
\label{sec:atom_existence}
\vspace{-1.0em}

\begin{wraptable}{r}{0.4\textwidth}
\vspace{-0.8em}
\centering
\footnotesize
\setlength{\tabcolsep}{2.8pt}
\renewcommand{\arraystretch}{1.08}
\begin{tabular}{lccccc}
\toprule
\textbf{Model}
& \multicolumn{2}{c}{\textbf{Learned}}
& \multicolumn{3}{c}{\textbf{Baselines}} \\
\cmidrule(lr){2-3}\cmidrule(lr){4-6}
$\boldsymbol{(S/T\uparrow)}$
& \textbf{Global}
& \textbf{Local}
& \textbf{Recon}
& \textbf{PCA}
& \textbf{Random} \\
\midrule
Mistral    & 1.27 & \textbf{1.42} & 0.69 & 0.70 & 0.68 \\
LLaMA & 1.88 & \textbf{1.98} & 0.89 & 0.83 & 0.79 \\
Gemma    & 1.59 & \textbf{2.12} & 0.62 & 0.65 & 0.59 \\
Qwen    & 1.63 & \textbf{1.84} & 0.73 & 0.76 & 0.71 \\
GLM      & 2.05 & \textbf{2.19} & 1.02 & 0.93 & 0.86 \\
DeepSeek   & 1.05 & \textbf{1.26} & 0.58 & 0.58 & 0.57 \\
Falcon    & 1.68 & \textbf{1.79} & 0.93 & 0.93 & 0.84 \\
\bottomrule
\end{tabular}
\vspace{-0.75em}
\caption{Semantic selectivity of learned global and locally reweighted atom frames compared with reconstruction-only and linear baselines. Local context improves $S/T$ across all seven models.}
\label{tab:atom_existence}
\vspace{-1.5em}
\end{wraptable}

The Invariant Atom Hypothesis predicts more than a low-dimensional semantic subspace: semantic motion should admit sparse coordinates that preferentially capture semantic change, remain stable to SP variation, and exhibit reproducible directional structure. We test these properties on held-out SP/SC transformations.

\vspace{-0.5em}
\noindent \textbf{Semantic selectivity requires invariance-aware learning.}
For an atom frame $A$, let
$e_{\mathrm{sem}}^{(n)}=\|A^\top\Delta z_{\mathrm{sem}}^{(n)}\|_2^2$
and let $\bar e_{\mathrm{nui}}(g)$ be the mean SP projection energy in neighborhood $g$. We define
\(
\mathrm{S/T}
=
1/N\sum_{n=1}^{N}
e_{\mathrm{sem}}^{(n)}
\space  / \space
\bar e_{\mathrm{nui}}(g(n)),
\)
so each semantic displacement is normalized by the nuisance response of its own neighborhood. 
$\mathrm{S/T}>1$ indicates preferential response to semantic rather than nuisance motion. 
Table~\ref{tab:atom_existence} shows $S/T>1$ for the learned frame across all seven models, while PCA and Random remain below one. 
For the locally conditioned frame, S/T is computed using 
$A(x_{\mathrm{anchor}})$ for both SC and SP projections around each anchor, 
so that selectivity is measured with respect to a single local coordinate 
system.
Reconstruction quality is evaluated by
$E_{\mathrm{rec}}
=\|\Delta z_{\mathrm{sem}}-\widehat{\Delta z}_{\mathrm{sem}}\|_2
/\|\Delta z_{\mathrm{sem}}\|_2$,
averaged over held-out SC samples. 

\vspace{-0.5em}
\noindent \textbf{SAE-style sparse reconstruction does not recover invariant coordinates.}
Recon-Only provides an explicit SAE-style sparse reconstruction baseline: it uses the same atom frame and sparse encoder but optimizes only $L_{\mathrm{sem}}$, reconstructing semantic displacement without SP-based invariance constraints. Despite equal or better reconstruction, its selectivity drops sharply, e.g., $1.27\!\rightarrow\!0.69$ on Mistral, $1.63\!\rightarrow\!0.73$ on Qwen, and $1.59\!\rightarrow\!0.62$ on Gemma. Coordinate drift likewise increases on Mistral and Qwen from $1.49$ to $2.34$ and from $1.77$ to $3.21$. Thus, sparse reconstruction alone does not yield invariant coordinates; transformation-aware constraints are necessary for nuisance suppression. Full objective ablations are reported in Appendix~\ref{app:atom_ablation}.

\vspace{-0.5em}
\noindent \textbf{Local context improves the relevance of shared atom directions.}
We further allow context-dependent diagonal reweighting,
\(
A(x)=A_0\operatorname{diag}(1+r(x)),
\)
which preserves atom directions while adapting their importance to the current semantic neighborhood. Local improves semantic reconstruction and $S/T$ across all seven models: for example, $S/T$ increases from $1.27$ to $1.42$ on Mistral, $1.59$ to $2.12$ on Gemma, and $1.05$ to $1.26$ on DeepSeek. Thus, the evidence supports a global-to-local organization: $A_0$ provides reusable semantic directions, while local context modulates their relevance without redefining the coordinate system. Full cross-model and directional-adaptation analyses appear in Appendix~\ref{app:global_local}.

\vspace{-0.5em}
\noindent \textbf{Semantic changes admit sparse compositional coordinates.}
We retrain the frame for
\(
k\in\{4,8,16,24,32,48,64\}.
\)
Increasing $k$ improves reconstruction, but semantic selectivity remains robust over a broad range. 
LLaMA maintains $\mathrm{S/T} > 1$ at every tested sparsity, and Gemma
remains at or above $1.58$ even at $k = 4$. We use $k = 32$ by default.
Representative sweeps are reported in Appendix~\ref{app:atom_sparsity}.
These results support a compositional representation in which each semantic change activates only a small subset of the shared atom frame.

\vspace{-0.5em}
\noindent \textbf{Atom coordinates are reproducible beyond subspace stability.}
To test whether learning repeatedly recovers similar directions rather than merely similar subspaces, we train five runs with different optimization seeds under a common PCA initialization and align atoms across runs by Hungarian matching on absolute cosine similarity. Sharing the initialization isolates optimization stochasticity, while atom-wise matching tests whether individual preferred directions recur despite permutation and sign ambiguity. Across all seven model families, the mean cosine between matched atoms from different learned runs ranges from $0.659$ to $0.770$, with subspace similarity from $0.809$ to $0.890$; both substantially exceed comparisons of learned atoms with the initial PCA basis or with random directions. Atom-level recovery is substantial but incomplete: the fraction of matched atoms exceeding $0.9$ cosine ranges from $23.9\%$ on Gemma to $46.8\%$ on Falcon. We therefore do not claim strict identifiability. Rather, the results support a stable semantic region containing nontrivial preferred directional coordinates, consistent with the proposed Invariant Atom Hypothesis. Full cross-model results are reported in Appendix~\ref{app:atom_reproducibility}.



\vspace{-1.25em}
\subsection{Causal Validation of Invariant Atoms}
\label{sec:causal}
\vspace{-1.0em}

\begin{wraptable}{r}{0.40\textwidth}
\vspace{-0.8em}
\centering
\footnotesize
\setlength{\tabcolsep}{3.1pt}
\renewcommand{\arraystretch}{1.08}
\begin{tabular}{lcccc}
\toprule
& \multicolumn{3}{c}{\textbf{Deletion}} & \textbf{Injection} \\
\cmidrule(lr){2-4}\cmidrule(l){5-5}
\textbf{Model}
& \textbf{Active}
& \textbf{Inactive}
& \textbf{Shuffled}
& \textbf{Active} \\
\midrule
Mistral    & 0.125 & 0.000 & $-$0.006 & 0.105 \\
LLaMA & 0.099 & $-$0.003 & $-$0.005 & 0.141 \\
Gemma    & 0.101 & 0.005 & 0.008 & 0.084 \\
Qwen    & 0.177 & 0.003 & $-$0.004 & 0.152 \\
GLM      & 0.074 & 0.001 & $-$0.005 & 0.084 \\
DeepSeek  & 0.290 & 0.002 & $-$0.013 & 0.284 \\
Falcon    & 0.169 & $-$0.007 & $-$0.003 & 0.145 \\
\bottomrule
\end{tabular}
\vspace{-0.75em}
\caption{Causal interventions ($m=32$), reported as median $R_{\mathrm{causal}}$. Deletion removes active atom contributions from an SC state; injection adds them to the anchor. Inactive atoms are norm-matched controls, while Shuffled preserves directions but disrupts learned coefficients. Positive values indicate movement toward the target distribution.}
\label{tab:causal_summary}
\vspace{-1.5em}
\end{wraptable}

The preceding results show that invariant atoms reconstruct semantic displacement while suppressing nuisance variation. We next ask whether they also functionally control model behavior. On $500$ held-out anchor--SC pairs, the semantic displacement
\(
\Delta z_{\mathrm{sem}}=z_{\mathrm{SC}}-z_{\mathrm{anchor}}
\)
is sparsely decomposed as
\(
\Delta z_{\mathrm{sem}}\approx\sum_i\alpha_i a_i.
\)
We rank atoms by $|\alpha_i|\|a_i\|$ and define the top-$m$ contribution
\(
\delta_m=\sum_{i\in\mathcal{I}_m}\alpha_i a_i.
\)
We intervene on the final-token hidden state at the learned layer and propagate the modified representation through the remaining network. Deletion uses
\(
z_{\mathrm{SC}}^{\mathrm{del}}=z_{\mathrm{SC}}-\delta_m
\)
to move the SC state toward the anchor, while injection uses
\(
z_{\mathrm{anchor}}^{\mathrm{inj}}=z_{\mathrm{anchor}}+\delta_m
\)
to move the anchor toward the SC state. Thus, the same sparse code is tested bidirectionally. 
We measure progress toward the target distribution by
\(
R_{\mathrm{causal}}
=
1-
D_{\mathrm{KL}}\!\left(p_{\mathrm{target}}\|p_{\mathrm{int}}\right)
/
D_{\mathrm{KL}}\!\left(p_{\mathrm{target}}\|p_{\mathrm{base}}\right),
\)
where $R_{\mathrm{causal}}>0$ indicates movement toward the target. KL is computed over the top $100$ logits. To control for generic perturbation magnitude, Inactive uses norm-matched atoms outside active top $32$, while Shuffled preserves selected directions but permutes and sign flips their coefficients.

\vspace{-0.5em}
\noindent \textbf{Identified atoms functionally control semantic information.}
Table~\ref{tab:causal_summary} shows consistent causal separation across all seven model families. Removing active atoms moves SC predictions toward the anchor, while inactive and shuffled controls remain near zero. Conversely, injecting the learned code moves anchor predictions toward the corresponding SC state. The active effect also increases with $m$ from $1$ to $32$, whereas the controls remain near zero. For DeepSeek-MoE, median deletion reaches $0.290$, compared with $0.002$ for inactive atoms and $-0.013$ for shuffled codes; Falcon3 shows a similarly clear separation despite heavier-tailed KL ratios. Full dose-response curves, Local-frame interventions, per-SC-type results, and distributional statistics appear in Appendix~\ref{app:causal}. 
The locally scaled frame also produces positive bidirectional effects across all seven models; thus, local reweighting preserves the causal meaning of the underlying atoms without replacing their shared directional organization. 
These results show that invariant atoms are not merely correlational coordinates of semantic displacement: their learned directions and coefficients causally influence model predictions. 



\vspace{-1.25em}
\subsection{Generalization of Invariant Atoms}
\label{sec:generalization}
\vspace{-0.8em}

The standard split evaluates unseen transformations within known semantic neighborhoods. We next test whether the learned coordinates remain useful when semantic content or nuisance transformations themselves are unseen during training.

\vspace{-0.5em}
\noindent \textbf{Atom coordinates transfer to unseen semantic neighborhoods.}
We partition the $1{,}400$ anchor-centered groups into $1{,}116$ training and $284$ test groups, withholding each test anchor and all of its SP/SC variants from atom learning; PCA initialization is also fit only on training groups. Across architectures, reconstruction transfers strongly: cosine similarity on unseen groups retains approximately $95$--$98\%$ of seen-group performance for both Global and Local. Semantic selectivity also transfers in most cases, with unseen $S/T$ ranging from $0.96$ to $1.76$ for Global and $0.94$ to $1.65$ for Local; DeepSeek is the main boundary case, remaining close to but slightly below one. Thus, the learned frame largely preserves its semantic structure on entirely unseen neighborhoods rather than depending on previously observed anchors. Full per-model results are reported in Appendix~\ref{app:generalization}.

\begin{wraptable}{r}{0.525\textwidth}
\vspace{-0.8em}
\centering
\footnotesize
\setlength{\tabcolsep}{3.0pt}
\renewcommand{\arraystretch}{1.08}
\begin{tabular}{lcc}
\toprule
\textbf{Generalization Test}
& \textbf{Global}
& \textbf{Local} \\
\midrule
Unseen neighborhoods: Cos transfer
& $0.95$--$0.98\times$
& $0.95$--$0.98\times$ \\
Unseen neighborhoods: $S/T$
& $0.96$--$1.76$
& $0.94$--$1.65$ \\
Held-out SP families: $S/T$
& $1.13$--$3.32$
& $1.75$--$5.79$ \\
\bottomrule
\end{tabular}
\vspace{-0.5em}
\caption{Generalization beyond the standard within-neighborhood split. Ranges are across evaluated model families and, for held-out SP families, all nuisance family splits.}
\label{tab:generalization_summary}
\vspace{-1.0em}
\end{wraptable}

\vspace{-0.5em}
\noindent \textbf{Invariance extends to unseen nuisance families.}
We further withhold two of the eight SP transformation families from training, using three splits: question/passive, casual/technical, and simplified/verbose. Across all seven model families and all three splits, held-out $S/T$ remains above one for both Global and Local, ranging from $1.13$--$3.32$ and $1.75$--$5.79$, respectively. Transfer is strongest for question/passive and weakest for simplified/verbose, indicating that invariance extends beyond the nuisance transformations used to learn the frame while remaining sensitive to distributional difficulty. Full results and an additional cross-generator transfer test are provided in Appendix~\ref{app:generalization}.



\vspace{-1.25em}
\subsection{Functional Validation of Atom Signatures}
\label{sec:functional_validation}
\vspace{-1.0em}

We use two lightweight downstream probes to test whether the learned coordinates remain functionally meaningful beyond the geometric objectives used to train them. These experiments are intended as validation of the representation rather than as standalone retrieval or adaptation benchmarks.

\begin{wraptable}{r}{0.30\textwidth}
\vspace{-0.8em}
\centering
\footnotesize
\setlength{\tabcolsep}{4.0pt}
\renewcommand{\arraystretch}{1.07}
\begin{tabular}{lccc}
\toprule
\textbf{Model}
& \textbf{PCA}
& \textbf{Global}
& \textbf{Local} \\
\midrule
Mistral  & .292 & \textbf{.458} & .370 \\
LLaMA    & .294 & \textbf{.403} & .309 \\
Gemma    & .242 & \textbf{.345} & .230 \\
Qwen     & .204 & \textbf{.291} & .228 \\
GLM      & .286 & \textbf{.363} & .288 \\
DeepSeek & .248 & \textbf{.368} & .260 \\
Falcon   & .258 & \textbf{.314} & .241 \\
\bottomrule
\end{tabular}
\vspace{-0.5em}
\caption{R@1 semantic retrieval using $256$-dimensional signatures.}
\label{tab:retrieval_summary}
\vspace{-1.2em}
\end{wraptable}

\vspace{-0.5em}
\noindent \textbf{Semantic identity under surface variation.}
We construct a gallery of $1{,}400$ anchor signatures and use held-out SP variants as queries, retrieving the nearest anchor by cosine similarity. Retrieval is successful when a paraphrased query is matched back to the anchor from the same semantic group, so the task directly tests whether the signature preserves semantic identity despite surface-form changes. We compare equally compact $256$-dimensional PCA, Global $A_0^\top z(x)$, and Local $A(x)^\top z(x)$ signatures. Table~\ref{tab:retrieval_summary} shows that Global outperforms PCA in R@1 across all seven models, indicating that the learned coordinates preserve semantic identity better than a generic variance-based subspace.

\vspace{-0.5em}
Global outperforms Local for direct retrieval. This complements the decomposition results in Section~\ref{sec:atom_existence} and the distinction is geometric: Global represents every anchor and query in the same coordinate system $A_0$, whereas Local uses an input-dependent frame $A(x)$, so two semantically equivalent inputs may be expressed under different coordinate-wise scalings. This reduces direct comparability even when the local decomposition itself is more selective. Accordingly, Local is better suited to context-sensitive semantic decomposition, while Global is better suited to cross-example matching. Full retrieval metrics and coordinate-mismatch diagnostics are reported in Appendix~\ref{app:retrieval}.
We distinguish the dense atom-space signature $s(x) = A_0^\top z(x)$ from the 
sparse reconstruction code 
$\alpha(\Delta z) = \mathrm{TopK}(W \Delta z + b,\, k)$. Retrieval uses the 
dense signature; semantic decomposition uses the sparse code. 

\begin{wraptable}{r}{0.35\textwidth}
\vspace{-1.2em}
\centering
\footnotesize
\setlength{\tabcolsep}{3.2pt}
\renewcommand{\arraystretch}{1.07}
\begin{tabular}{lccc}
\toprule
\textbf{Model}
& \textbf{Base}
& \textbf{Fine-tuned}
& \textbf{Distilled} \\
\midrule
Mistral  & .849 & .852 & .859 \\
LLaMA    & .872 & .873 & .870 \\
Qwen     & .866 & .861 & .861 \\
GLM      & .869 & .866 & .866 \\
DeepSeek & .851 & .851 & .851 \\
Falcon   & .857 & .850 & .853 \\
Gemma    & .868 & .866 & .866 \\
\bottomrule
\end{tabular}
\vspace{-0.5em}
\caption{SC-type classification using atom signatures after modification.}
\label{tab:model_modification_summary}
\vspace{-1.0em}
\end{wraptable}

\vspace{-0.5em}
\noindent \textbf{Semantic stability under model modification.}
We next ask whether the semantic organization of atom coordinates survives changes to the underlying model parameters. For each architecture, we extract Global signatures from the base model and use them to train a seven-way classifier that predicts the SC transformation type (entity, attribute, relation, negation, quantity, temporal, or intent). We then apply this same classifier, without retraining, to signatures extracted from LoRA fine-tuned and distilled variants of the model. Thus, preservation of classification accuracy indicates that the semantic coordinate structure learned on the base model remains aligned after modification. 
Table~\ref{tab:model_modification_summary} shows that SC-type classification remains nearly unchanged across fine-tuned and distilled variants, indicating that the semantic organization encoded by the shared atom coordinates is largely preserved under moderate parameter adaptation. Full comparisons with Raw, PCA, Local, sparse TopK signatures, and signature-drift analyses are reported in Appendix~~\ref{app:model_modification}.

%% file: sections/suplementery.tex
\startcontents[appendix]
\section*{Appendix Contents}
\printcontents[appendix]{}{1}{}

\section{Additional Dataset Construction Details}
\label{app:data_details}

The invariant-atom framework relies on controlled local semantic neighborhoods that distinguish semantic-preserving (SP) variation from semantic-changing (SC) variation. This appendix provides additional details on the composition and construction of the dataset used throughout the experiments.

\paragraph{Dataset scale and semantic coverage.}
The primary dataset contains $1{,}400$ anchor-centered semantic groups and $64{,}093$ sentences in total. These groups cover $274$ distinct topics distributed across $13$ broad domains, including animals, science, countries, technology, food, history, mathematics, arts, health, philosophy, sports, environment, and a small pilot subset. The distribution is intentionally heterogeneous rather than uniform: individual topics contribute between one and thirteen anchor groups, with an average of approximately $5.1$ groups per topic. This construction exposes the atom-learning procedure to a broad range of semantic content while preserving the local neighborhood structure required by the formulation.

\begin{wraptable}{r}{0.35\textwidth}
\vspace{-1.0em}
\centering
\footnotesize
\setlength{\tabcolsep}{4pt}
\renewcommand{\arraystretch}{1.08}

\begin{tabular}{lc}
\toprule
\textbf{Domain} & \textbf{\# Anchor Groups} \\
\midrule
Animals      & 169 \\
Science      & 167 \\
Countries    & 165 \\
Technology   & 133 \\
Food         & 110 \\
History      & 108 \\
Mathematics  & 107 \\
Arts         & 90 \\
Health       & 89 \\
Philosophy   & 87 \\
Sports       & 85 \\
Environment  & 79 \\
Pilot data   & 11 \\
\midrule
\textbf{Total} & \textbf{1,400} \\
\bottomrule
\end{tabular}

\vspace{-0.75em}
\caption{Distribution of anchor neighborhoods across semantic domains.}
\label{tab:domain_distribution}
\vspace{-1.5em}
\end{wraptable}

\paragraph{Semantic-preserving transformations.}
Each anchor is associated with multiple SP variants that preserve propositional meaning while changing surface realization. Across the dataset, there are $33{,}335$ SP sentences spanning eight transformation families: formal, casual, question, technical, simplified, verbose, passive, and imperative. These perturbations alter lexical choice, syntactic form, grammatical construction, and style while maintaining the underlying semantic content. They therefore provide empirical samples of nuisance motion around each anchor.

For example, for the anchor
\emph{``How is battery storage characterized in renewable energy?''},
SP variants include
\emph{``In what manner is battery storage described within the context of renewable energy?''}
and
\emph{``What's the deal with battery storage in renewable energy?''}.
Although the surface forms differ substantially, the underlying query remains unchanged.

\paragraph{Semantic-changing transformations.}
The dataset contains $29{,}358$ SC sentences spanning seven controlled semantic-change types: entity, attribute, relation, negation, quantity, temporal, and intent. These transformations alter the semantic state while often retaining much of the lexical and syntactic structure of the anchor. This is important for separating semantic displacement from generic textual distance: a local edit can change meaning substantially even when most of the sentence remains fixed.

Using the same renewable-energy anchor, representative SC variants include
\emph{``How is capacitor storage characterized in renewable energy?''}
and
\emph{``How is solar panel storage characterized in renewable energy?''},
which modify the entity while preserving the surrounding structure. Similarly, for the anchor
\emph{``Describe the conservation of dolphins,''}
entity-level SC variants include
\emph{``Describe the conservation of whales''}
and
\emph{``Describe the conservation of seals.''}

\paragraph{Local neighborhood structure.}
Each semantic group is organized around a single anchor and contains approximately $24$ SP and $21$ SC variants on average. The corresponding representation differences
\[
\Delta z_{\mathrm{nui}}
=
\Phi_\ell(x^{\mathrm{SP}})-\Phi_\ell(x),
\qquad
\Delta z_{\mathrm{sem}}
=
\Phi_\ell(x^{\mathrm{SC}})-\Phi_\ell(x)
\]
provide empirical samples of nuisance and semantic motion, respectively. The atom-learning objective is therefore defined over local displacement vectors rather than absolute activations alone.

\paragraph{Training and evaluation split.}
For the standard experiments, the SP and SC variants within each group are divided using an interleaved even/odd split. Even-indexed variants are used for training, while odd-indexed variants are reserved for evaluation. 
This ensures that the learned atom frame, sparse encoder, and local
modulation functions are tested on transformations that are not used for
gradient updates.
Importantly, this split evaluates generalization to held-out transformations within known semantic neighborhoods; it does not by itself test transfer to entirely unseen anchors or topics. We therefore evaluate group-disjoint generalization separately in Section~\ref{sec:generalization}.

\paragraph{Manual quality verification.}
To verify the quality of the automatically generated perturbations, we manually inspected a sample of generated SP and SC variants across semantic groups and transformation types. The inspected SP examples preserved the meaning of their anchors while varying surface form, whereas the inspected SC examples introduced the intended semantic change without substantially altering unrelated content. This manual inspection was used as a quality check on the generation procedure rather than as an additional filtering or annotation stage.

\paragraph{Cross-generator perturbation set.}
The primary perturbation dataset is generated with Mistral-7B-Instruct-v0.3. To test whether the learned atom structure depends on a particular paraphrase generator, we additionally construct an independently generated perturbation set using Qwen2.5-7B-Instruct. These data are used exclusively for the cross-generator generalization experiments in Section~\ref{sec:generalization}.



\section{Additional Evidence for Invariant Atoms}
\label{app:atom_evidence}

This section provides additional evidence for Section~\ref{sec:atom_existence}, including complete cross-model baseline comparisons, semantic-change and nuisance-style breakdowns, sparsity sweeps, cross-seed reproducibility, objective ablations, and layer-wise behavior.

\subsection{Full Cross-Model Baseline Comparison}
\label{app:atom_baselines}

Table~\ref{tab:full_dictionary_comparison} expands the main comparison beyond $S/T$. Learned atoms reconstruct semantic displacement comparably to PCA while consistently providing stronger semantic--nuisance separation and generally lower coordinate drift.

\begin{table}[H]
\centering
\scriptsize
\setlength{\tabcolsep}{3.2pt}
\renewcommand{\arraystretch}{1.05}
\begin{tabular}{llccccc}
\toprule
\textbf{Model} & \textbf{Frame}
& $E_{\mathrm{rec}}\downarrow$
& $\mathrm{Cos}\uparrow$
& $S/T\uparrow$
& $D_{\mathrm{coord}}\downarrow$
& $\mathrm{EffNum}$ \\
\midrule
Mistral
& Learned & 0.701 & 0.703 & \textbf{1.27} & \textbf{1.49} & 21.5 \\
& PCA     & 0.716 & 0.690 & 0.70 & 2.25 & 25.0 \\
& Random  & 0.985 & 0.171 & 0.68 & 2.04 & 30.8 \\
\midrule
LLaMA
& Learned & 0.729 & 0.670 & \textbf{1.88} & \textbf{1.61} & 23.0 \\
& PCA     & 0.750 & 0.652 & 0.83 & 2.37 & 26.5 \\
& Random  & 0.984 & 0.176 & 0.79 & 2.06 & 30.8 \\
\midrule
Gemma
& Learned & 0.711 & 0.686 & \textbf{1.59} & \textbf{2.50} & 21.8 \\
& PCA     & 0.723 & 0.679 & 0.65 & 3.77 & 25.9 \\
& Random  & 0.982 & 0.188 & 0.59 & 3.34 & 30.8 \\
\midrule
Qwen
& Learned & 0.682 & 0.720 & \textbf{1.63} & \textbf{1.77} & 22.3 \\
& PCA     & 0.686 & 0.718 & 0.76 & 2.87 & 25.0 \\
& Random  & 0.982 & 0.187 & 0.71 & 2.51 & 30.8 \\
\midrule
GLM
& Learned & 0.716 & 0.683 & \textbf{2.05} & \textbf{1.48} & 22.9 \\
& PCA     & 0.734 & 0.668 & 0.93 & 2.06 & 25.7 \\
& Random  & 0.984 & 0.176 & 0.86 & 1.80 & 30.8 \\
\midrule
DeepSeek
& Learned & 0.691 & 0.712 & \textbf{1.05} & \textbf{1.84} & 23.0 \\
& PCA     & 0.698 & 0.707 & 0.58 & 2.77 & 25.4 \\
& Random  & 0.968 & 0.252 & 0.57 & 2.41 & 30.8 \\
\midrule
Falcon
& Learned & 0.767 & 0.625 & \textbf{1.68} & \textbf{1.45} & 23.6 \\
& PCA     & 0.787 & 0.606 & 0.93 & 1.95 & 26.7 \\
& Random  & 0.979 & 0.203 & 0.84 & 1.73 & 30.8 \\
\bottomrule
\end{tabular}
\caption{Full comparison across seven model families. Learned atoms consistently achieve $S/T>1$ while retaining competitive semantic reconstruction and generally lower coordinate drift than PCA. We report relative reconstruction error $E_{\mathrm{rec}} = \left\|{\Delta z_{\mathrm{sem}} - \Delta \hat{z}_{\mathrm{sem}}}\right\|_2 / \left\|{\Delta z_{\mathrm{sem}}}\right\|_2$, averaged over held-out SC samples. This measures normalized displacement recovery.}
\label{tab:full_dictionary_comparison}
\end{table}

We report two additional evaluation metrics throughout. Coordinate drift 
measures the instability of atom-space coordinates under semantic-preserving 
variation:
\[
  D_{\mathrm{coord}}
  = \frac{1}{N_{\mathrm{SC}}}\sum_{n=1}^{N_{\mathrm{SC}}}
    \left[
      \frac{1}{M'}\sum_{m=1}^{M'}
      \frac{\big\| A^{\top}(z^{\mathrm{SC}}_n - z)
                 - A^{\top}(z^{\mathrm{SC}}_n - z^{\mathrm{SP}}_m) \big\|_2^2}
           {\big\| A^{\top}(z^{\mathrm{SC}}_n - z) \big\|_2^2}
    \right],
  \qquad M' = \min(5, M),
\]
where $z$ is the anchor, $z^{\mathrm{SC}}_n$ is the $n$-th held-out SC
variant, and $z^{\mathrm{SP}}_m$ is the $m$-th held-out SP variant; drift is
computed against the first $M'$ held-out SP variants of each anchor. 

For the Local model, both terms use the anchor-conditioned frame
$A=A(x)$. 
Lower values indicate more 
stable atom coordinates under paraphrasing. The effective number of active 
atoms is the inverse Herfindahl index of coefficient magnitudes:

\[
  \mathrm{EffNum}
  = \frac{\big(\sum_{j=1}^{K} |\alpha_j|\big)^2}
         {\sum_{j=1}^{K} |\alpha_j|^2}.
\]
If all weight concentrates on a single atom, $\mathrm{EffNum} = 1$; if weight 
is spread uniformly across $k$ atoms, $\mathrm{EffNum} = k$.

PCA captures high-variance directions but remains below $S/T=1$ across architectures, while random frames reconstruct poorly. Thus, the learned selectivity is not explained by dimensionality reduction or variance capture alone.

\paragraph{SAE-style reconstruction controls across all models.}
We additionally train Recon-Only on every architecture. This control uses the same atom frame and sparse encoder as the full model but optimizes only semantic reconstruction, providing an SAE-style baseline without transformation-aware invariance supervision.

\begin{table}[H]
\centering
\scriptsize
\setlength{\tabcolsep}{3.3pt}
\renewcommand{\arraystretch}{1.05}
\begin{tabular}{lcccccc}
\toprule
& \multicolumn{2}{c}{$E_{\mathrm{rec}}\downarrow$}
& \multicolumn{2}{c}{$S/T\uparrow$}
& \multicolumn{2}{c}{$D_{\mathrm{coord}}\downarrow$} \\
\cmidrule(lr){2-3}\cmidrule(lr){4-5}\cmidrule(lr){6-7}
\textbf{Model}
& \textbf{Full} & \textbf{Recon}
& \textbf{Full} & \textbf{Recon}
& \textbf{Full} & \textbf{Recon} \\
\midrule
Mistral   & 0.701 & \textbf{0.671} & \textbf{1.27} & 0.69 & \textbf{1.49} & 2.34 \\
LLaMA     & 0.729 & \textbf{0.709} & \textbf{1.88} & 0.89 & \textbf{1.61} & 2.55 \\
Gemma     & 0.711 & \textbf{0.676} & \textbf{1.59} & 0.62 & \textbf{2.50} & 4.14 \\
Qwen      & 0.682 & \textbf{0.652} & \textbf{1.63} & 0.73 & \textbf{1.77} & 3.21 \\
GLM       & 0.716 & \textbf{0.695} & \textbf{2.05} & 1.02 & \textbf{1.48} & 2.13 \\
DeepSeek  & 0.691 & \textbf{0.671} & \textbf{1.05} & 0.58 & \textbf{1.84} & 2.86 \\
Falcon    & 0.767 & \textbf{0.758} & \textbf{1.68} & 0.93 & \textbf{1.45} & 2.11 \\
\bottomrule
\end{tabular}
\caption{Full versus SAE-style Recon-Only training across architectures. Reconstruction-only learning often improves $E_{\mathrm{rec}}$ but substantially reduces semantic selectivity and increases coordinate drift. }
\label{tab:recon_all_models}
\end{table}

The pattern is consistent across all seven architectures: Recon-Only typically reconstructs semantic displacement better than the full model, yet its $S/T$ falls toward or below one and coordinate drift increases substantially. The contrast is especially strong for Gemma ($1.59\!\rightarrow\!0.62$), LLaMA ($1.88\!\rightarrow\!0.89$), and Qwen ($1.63\!\rightarrow\!0.73$). Thus, sparse reconstruction and semantic invariance are distinct objectives: accurate reconstruction alone does not organize semantic motion into stable nuisance-suppressing coordinates.

\subsection{Selectivity Across Semantic-Change Types}
\label{app:atom_sc_types}

The aggregate $S/T$ score averages across heterogeneous semantic changes. To test whether selectivity is concentrated in a single perturbation type, we evaluate each of the seven SC categories separately. Table~\ref{tab:sc_type_examples} reports representative results for LLaMA, Qwen, GLM, and DeepSeek; the same qualitative pattern is observed across the remaining models.
Per-type values use the same definition as the aggregate S/T, restricted
to SC samples of that type; the aggregate averages over all held-out SC
samples and is therefore weighted by the number of samples of each type.

Selectivity varies systematically with semantic-change type. Negation and temporal changes produce the strongest separation, while entity and attribute substitutions are harder. Negation and temporal operators often alter a discrete semantic relation while preserving most surface form, whereas entity substitutions can induce changes that overlap with lexical nuisance variation. The aggregate learned advantage is therefore not driven by a single SC category.

\begin{table}[H]
\centering
\small
\setlength{\tabcolsep}{5pt}
\renewcommand{\arraystretch}{1.05}
\begin{tabular}{lcccc}
\toprule
\textbf{SC Type}
& \textbf{LLaMA}
& \textbf{Qwen}
& \textbf{GLM}
& \textbf{DeepSeek} \\
\midrule
Entity    & 0.56 & 0.56 & 0.58 & 0.49 \\
Attribute & 0.82 & 0.74 & 0.83 & 0.60 \\
Relation  & 1.35 & 1.15 & 1.27 & 0.81 \\
Negation  & 3.80 & 3.57 & 4.82 & 1.69 \\
Quantity  & 1.52 & 1.53 & 1.76 & 0.91 \\
Temporal  & 2.32 & 1.66 & 2.05 & 1.46 \\
Intent    & 1.57 & 1.21 & 1.56 & 0.84 \\
\midrule
All       & 1.88 & 1.63 & 2.05 & 1.05 \\
\bottomrule
\end{tabular}
\caption{$S/T$ of the learned atom frame by semantic-change type. Logical and temporal changes are generally most separable, whereas entity and attribute substitutions are more difficult because they overlap more strongly with lexical variation.}
\label{tab:sc_type_examples}
\end{table}

\subsection{Nuisance Suppression Across SP Styles}
\label{app:atom_sp_styles}

We analyze the atom-response energy
$\|A^\top\Delta z_{\mathrm{nui}}\|_2^2$
for each SP transformation family. Table~\ref{tab:sp_style_qwen} shows lower responses for the learned frame than for PCA and Random across all eight styles. Because this quantity depends on atom magnitudes, the absolute reductions measure suppression in the learned representation and do not, by themselves, establish greater angular separation
from nuisance directions.

\begin{table}[H]
\centering
\small
\setlength{\tabcolsep}{5pt}
\renewcommand{\arraystretch}{1.05}
\begin{tabular}{lccc}
\toprule
\textbf{SP Style}
& \textbf{Learned}
& \textbf{PCA}
& \textbf{Random} \\
\midrule
Formal     & 7.98  & 2257 & 208 \\
Casual     & 10.55 & 3168 & 340 \\
Question   & 7.13  & 1965 & 194 \\
Technical  & 10.18 & 2599 & 256 \\
Simplified & 10.90 & 2889 & 292 \\
Verbose    & 13.07 & 4897 & 465 \\
Passive    & 14.36 & 4870 & 522 \\
Imperative & 11.85 & 3196 & 325 \\
\bottomrule
\end{tabular}
\caption{Nuisance projection energy across SP transformation families for Qwen2.5-7B. Lower is better. Learned atoms suppress nuisance motion across all eight styles rather than specializing to a particular paraphrase template.}
\label{tab:sp_style_qwen}
\end{table}

Passive, verbose, and imperative transformations tend to induce larger nuisance shifts, but suppression remains strong across all eight styles. This supports the interpretation that the learned frame captures a broad family of semantic-preserving variation rather than a single paraphrase mechanism.

\subsection{Full Sparsity Analysis}
\label{app:atom_sparsity}

The main paper reports that atom coordinates remain semantically selective over a broad range of sparsity levels. Here we provide complete sweeps for representative model families. For each $k$, the atom frame is retrained independently under the same optimization protocol.

Increasing $k$ improves reconstruction, as expected, while $S/T$ remains comparatively stable. LLaMA stays between $1.68$ and $1.91$ across the full sweep, and Gemma remains above $1.58$ even at $k=4$. DeepSeek is more demanding: its $S/T$ rises from below one at very small support to above one once a moderately larger support is allowed. Thus, the precise sparsity threshold is model-dependent, but semantic selectivity does not require dense use of the atom frame.

At the default $k=32$, the effective support is typically only about $20$--$24$ atoms across architectures. This motivates $k=32$ as a conservative operating point that balances reconstruction with sparse semantic composition.

\begin{table}[H]
\centering
\small
\setlength{\tabcolsep}{4pt}
\renewcommand{\arraystretch}{1.05}
\begin{tabular}{cccccc}
\toprule
$k$
& $E_{\mathrm{rec}}\downarrow$
& $\mathrm{Cos}\uparrow$
& $S/T\uparrow$
& $D_{\mathrm{coord}}\downarrow$
& $\mathrm{EffNum}$ \\
\midrule
\multicolumn{6}{c}{\textbf{LLaMA-3.1-8B}} \\
\midrule
4  & 0.918 & 0.361 & 1.88 & 2.38 & 3.2 \\
8  & 0.875 & 0.462 & 1.83 & 2.28 & 6.2 \\
16 & 0.829 & 0.542 & 1.68 & 2.02 & 11.8 \\
24 & 0.794 & 0.592 & 1.91 & 1.83 & 17.2 \\
32 & 0.772 & 0.622 & 1.86 & 1.72 & 22.1 \\
48 & 0.736 & 0.664 & 1.84 & 1.59 & 31.4 \\
64 & 0.715 & 0.688 & 1.85 & 1.52 & 40.6 \\
\midrule
\multicolumn{6}{c}{\textbf{Gemma-2-9B}} \\
\midrule
4  & 0.893 & 0.418 & 1.58 & 3.00 & 3.1 \\
8  & 0.856 & 0.491 & 1.62 & 2.89 & 6.0 \\
16 & 0.796 & 0.582 & 1.88 & 2.73 & 11.0 \\
24 & 0.759 & 0.631 & 1.84 & 2.52 & 16.5 \\
32 & 0.739 & 0.657 & 1.81 & 2.45 & 20.97 \\
48 & 0.711 & 0.689 & 1.80 & 2.30 & 30.3 \\
64 & 0.695 & 0.706 & 1.78 & 2.24 & 38.8 \\
\bottomrule
\end{tabular}
\caption{Representative sparsity sweeps. Reconstruction improves as more atoms are available, while $S/T$ remains stable over a broad range. Effective support remains substantially below the full set of 256 atoms. Sparsity-sweep models are independently trained with random orthonormal 
initialization; the $k{=}32$ entry may therefore differ from the corresponding row in 
Table~\ref{tab:full_dictionary_comparison}, which uses PCA initialization.}
\label{tab:sparsity_full}
\end{table}

\subsection{Cross-Seed Reproducibility of Atom Coordinates}
\label{app:atom_reproducibility}

We test whether atom learning repeatedly recovers similar semantic directions under optimization stochasticity. For each of the seven model families, we train five runs using seeds
\(
\{42,137,256,512,1024\}.
\)
All runs for a given model start from the same PCA initialization. Consequently, differences across runs arise from stochastic optimization rather than from different initial subspaces. This experiment is deliberately stricter than measuring only subspace overlap: two runs may span nearly the same semantic region while representing that region with arbitrarily rotated bases. Reproducibility at the level of individual atoms therefore provides evidence for preferred directions within the recovered semantic region.

Atom indices and signs are not intrinsically ordered, so two equivalent solutions need not assign the same semantic direction to the same column or sign. We therefore align every pair of learned frames using Hungarian matching on absolute inner products between column-normalized atoms 
(equivalent to absolute cosine similarity for unit-norm columns). 
For frames $A^{(s)}$ and $A^{(t)}$, we solve
\[
\pi^\star
=
\arg\max_{\pi}
\frac{1}{K}
\sum_{i=1}^{K}
\left|
\left(a_i^{(s)}\right)^\top
a_{\pi(i)}^{(t)}
\right|.
\]
With five runs, this yields ten distinct learned--learned pairwise comparisons per model.

We report four complementary quantities. \emph{Mean $|\cos|$} is the average absolute cosine similarity after optimal atom matching and measures overall directional agreement. \emph{Median} is less sensitive to a small number of exceptionally stable or unstable atoms. \emph{$\%>0.9$} reports the fraction of matched atoms whose absolute cosine exceeds $0.9$, providing a stricter measure of near-recovery at the individual-direction level. \emph{Subspace} measures similarity between the spans of the two frames using principal-angle similarity and therefore ignores basis rotation within the recovered semantic region.

Two references help interpret these values. \emph{Learned--PCA} aligns each learned frame with the common PCA initialization and measures how much atom-level agreement can be attributed to remaining close to the initialization. \emph{Learned--Random} compares the learned frame with a random orthogonal frame and provides a chance-level reference. High learned--learned agreement relative to both controls therefore indicates that optimization repeatedly converges toward similar directions rather than merely inheriting the PCA basis or exhibiting accidental alignment.

The pattern is consistent across architectures. Mean matched-atom cosine between independently optimized runs ranges from $0.659$ on Gemma to $0.770$ on Falcon, whereas alignment with the PCA initialization is only $0.287$--$0.340$ and alignment with random directions is $0.045$--$0.065$. 

The same distinction appears at the subspace level. Learned--learned subspace similarity ranges from $0.809$ to $0.890$ across all seven models, confirming that optimization repeatedly recovers a highly similar semantic region. Individual atom recovery is weaker but still substantial: median matched cosine ranges from $0.688$ to $0.886$, and between $23.9\%$ and $46.8\%$ of atoms exceed $0.9$ cosine similarity. Hence, the recovered object is more structured than an arbitrary basis of a stable subspace, but not every atom is uniquely determined.

These findings provide evidence for reproducible directional structure under optimization stochasticity with a shared initialization. The reproducibility exceeds what would be expected from trivially remaining near the PCA basis, but does not establish that the same directions would be recovered from independent initializations. We therefore interpret the results as consistent with preferred directions within the learned semantic region, without claiming strict identifiability. 
The empirical picture is instead a reproducible semantic region containing a nontrivial set of preferred directions together with less stable coordinates, which is the level of structure required by the Invariant Atom Hypothesis.
Subspace similarity is computed as the mean cosine of the principal angles 
between the two $K$-dimensional subspaces spanned by each pair of learned 
frames.

\begin{table}[H]
\centering
\small
\setlength{\tabcolsep}{4.2pt}
\renewcommand{\arraystretch}{1.06}
\begin{tabular}{llcccc}
\toprule
\textbf{Model}
& \textbf{Comparison}
& \textbf{Mean $|\cos|$}
& \textbf{Median}
& $\mathbf{\%>0.9}$
& \textbf{Subspace} \\
\midrule
Mistral
& Learned--Learned & \textbf{0.711} & \textbf{0.820} & \textbf{36.4} & \textbf{0.845} \\
& Learned--PCA     & 0.287 & 0.161    & 0.4   & 0.656 \\
& Learned--Random  & 0.045 & 0.044    & 0.0  & 0.213 \\
\midrule
LLaMA
& Learned--Learned & \textbf{0.747} & \textbf{0.869} & \textbf{43.8} & \textbf{0.867} \\
& Learned--PCA     & 0.300 & 0.159 & 1.0  & 0.701 \\
& Learned--Random  & 0.045 & 0.045 & 0.0  & 0.213 \\
\midrule
Gemma
& Learned--Learned & \textbf{0.659} & \textbf{0.688} & \textbf{23.9} & \textbf{0.878} \\
& Learned--PCA     & 0.300 & 0.193 & 0.2  & 0.686 \\
& Learned--Random  & 0.049 & 0.049 & 0.0  & 0.228 \\
\midrule
Qwen
& Learned--Learned & \textbf{0.660} & \textbf{0.766} & \textbf{26.7} & \textbf{0.809} \\
& Learned--PCA     & 0.298 & 0.178 & 0.1  & 0.621 \\
& Learned--Random  & 0.048 & 0.048 & 0.0  & 0.228 \\
\midrule
GLM
& Learned--Learned & \textbf{0.722} & \textbf{0.849} & \textbf{41.1} & \textbf{0.850} \\
& Learned--PCA     & 0.287 & 0.158 & 0.6  & 0.684 \\
& Learned--Random  & 0.046 & 0.046 & 0.0  & 0.214 \\
\midrule
DeepSeek
& Learned--Learned & \textbf{0.699} & \textbf{0.781} & \textbf{34.1} & \textbf{0.855} \\
& Learned--PCA     & 0.319 & 0.186 & 2.3  & 0.705 \\
& Learned--Random  & 0.065 & 0.064 & 0.0  & 0.305 \\
\midrule
Falcon
& Learned--Learned & \textbf{0.770} & \textbf{0.886} & \textbf{46.8} & \textbf{0.890} \\
& Learned--PCA     & 0.340 & 0.183 & 1.7  & 0.735 \\
& Learned--Random  & 0.052 & 0.052 & 0.0  & 0.247 \\
\bottomrule
\end{tabular}
\caption{Cross-seed reproducibility of the global atom frame after optimal permutation and sign alignment. Each model is trained five times from a common PCA initialization, yielding ten learned--learned comparisons. Learned--PCA measures residual alignment with the initialization, while Learned--Random provides a chance-level reference. Across all seven models, independently optimized frames recover both highly similar semantic subspaces and substantially aligned individual atom directions.}
\label{tab:atom_seed_full}
\end{table}

\subsection{Objective Ablations}
\label{app:atom_ablation}

The all-model Recon-Only comparison in Table~\ref{tab:recon_all_models} establishes that sparse reconstruction alone is insufficient for invariant coordinates. We next use Mistral and Qwen as representative models to isolate which transformation-aware objectives produce this separation. Starting from the full model, we remove the nuisance-suppression, coordinate-stability, and orthogonality terms individually and jointly. Checkpoint selection follows the objective of each ablation; in particular, Recon-Only checkpoints are selected by reconstruction quality rather than an invariance-aware criterion.

\begin{table}[H]
\centering
\small
\setlength{\tabcolsep}{4.2pt}
\renewcommand{\arraystretch}{1.07}
\begin{tabular}{lcccc}
\toprule
& \multicolumn{2}{c}{$S/T\uparrow$}
& \multicolumn{2}{c}{$D_{\mathrm{coord}}\downarrow$} \\
\cmidrule(lr){2-3}\cmidrule(lr){4-5}
\textbf{Objective}
& \textbf{Mistral}
& \textbf{Qwen}
& \textbf{Mistral}
& \textbf{Qwen} \\
\midrule
Full             & \textbf{1.27} & \textbf{1.63} & \textbf{1.49} & \textbf{1.77} \\
Recon-Only       & 0.69 & 0.73 & 2.34 & 3.21 \\
No-$L_{\mathrm{nui}}$   & 1.24 & 1.54 & 1.51 & 1.92 \\
No-$L_{\mathrm{stress}}$& 1.26 & 1.58 & 1.50 & 1.81 \\
No-$L_{\mathrm{nui}}$/$L_{\mathrm{stress}}$
                 & 0.69 & 0.74 & 2.40 & 3.25 \\
No-$L_{\mathrm{orth}}$  & 1.28 & 1.63 & 1.49 & 1.77 \\
\bottomrule
\end{tabular}
\caption{Objective ablations on Mistral and Qwen. Removing both transformation-aware invariance terms reproduces the Recon-Only regime, whereas removing either term alone causes only moderate degradation.}
\label{tab:objective_ablation}
\end{table}

Removing either $L_{\mathrm{nui}}$ or $L_{\mathrm{stress}}$ alone causes only moderate degradation, whereas removing both nearly reproduces Recon-Only. Under the implemented formulation, $L_{\mathrm{stress}}=L_{\mathrm{nui}}$ for both Global and Local, so these ablations primarily vary the effective strength of nuisance suppression: removing one term retains a reduced nuisance penalty, while removing both eliminates it.

\subsection{Layer-Wise Emergence}
\label{app:atom_layers}

Finally, we examine how invariant atom structure varies through transformer depth. A separate atom frame is trained at each sampled layer using the same SP/SC construction,
with random initialization and optimization protocol. Semantic selectivity is not confined to a single manually chosen layer, although its strength varies with depth.

For LLaMA, $S/T$ is already $1.83$ at layer 4, reaches $1.91$ at layer 8, and remains above one through the final layer. Coordinate drift decreases from $2.82$ at layer 4 to approximately $1.4$--$1.5$ in later layers. Falcon exhibits a related pattern: selectivity rises from $1.31$ at layer 4 to $1.91$ at layer 12 before decreasing toward $1.05$ at layer 28, while coordinate stability improves with depth. Thus, semantic selectivity and coordinate stability are related but distinct properties of the representation hierarchy.

Some architectures exhibit unusually large early-layer $S/T$ when SP projection energy becomes very small. For example, Gemma reaches a nominal $S/T=14.03$ at layer 4 because SP coverage is unusually low. We therefore do not interpret the maximum ratio alone as evidence of superior semantic geometry and select primary layers jointly based on reconstruction, semantic selectivity, and coordinate stability.

Across architectures, invariant atom structure appears over substantial portions of transformer depth rather than at one isolated layer. Mid-layer representations often provide the strongest balance between reconstruction and semantic--nuisance separation, whereas deeper layers frequently exhibit more stable atom coordinates. This layer dependence is consistent with invariant atoms characterizing learned representation geometry rather than a fixed property of the input embedding space.



\section{Additional Causal Intervention Results}
\label{app:causal}

This section provides additional details and complete results for the causal validation in Section~\ref{sec:causal}. We report the full intervention dose response, comparisons between the global and locally scaled atom frames, and additional robustness analyses.

\subsection{Intervention Protocol}
\label{app:causal_protocol}

For each model, we sample $500$ held-out anchor--SC pairs and extract the final-token hidden states at the target layer used for atom learning. Given an anchor $x$ and semantic-changing variant $x^{\mathrm{SC}}$, we form
\[
\Delta z_{\mathrm{sem}}
=
z_{\mathrm{SC}}-z_{\mathrm{anchor}},
\qquad
\alpha = E(\Delta z_{\mathrm{sem}}),
\]
where $E$ denotes the learned sparse encoder. Active atoms are ranked by their contribution
\[
c_i = |\alpha_i|\,\|a_i\|_2,
\]
and the top-$m$ components define
\[
\delta_m
=
\sum_{i\in\mathcal{I}_m}\alpha_i a_i,
\qquad
m\in\{1,4,8,16,32\}.
\]
Deletion intervenes on the SC representation as
\[
z_{\mathrm{SC}}^{\mathrm{del}}
=
z_{\mathrm{SC}}-\delta_m,
\]
with the anchor distribution as the target, whereas injection uses
\[
z_{\mathrm{anchor}}^{\mathrm{inj}}
=
z_{\mathrm{anchor}}+\delta_m,
\]
with the SC distribution as the target. Modified hidden states are inserted through forward hooks and propagated through the remaining transformer layers.

We compare three intervention conditions. \emph{Active} uses the top-$m$ contributing atoms with their learned coefficients. \emph{Inactive} selects atoms outside the active top-$32$ set and rescales the resulting intervention to match the norm of the active intervention. \emph{Shuffled} uses the active atom directions but permutes and sign-flips their coefficients, preserving the intervention ingredients while destroying the learned atom--coefficient correspondence. These controls distinguish semantic effects associated with the learned sparse code from effects caused by perturbation magnitude or arbitrary directions within the atom frame.

For an intended target distribution $p_{\mathrm{target}}$, we quantify causal progress as
\[
R_{\mathrm{causal}}
=
1-
\frac{
D_{\mathrm{KL}}\!\left(
p_{\mathrm{target}}\|p_{\mathrm{int}}
\right)
}{
D_{\mathrm{KL}}\!\left(
p_{\mathrm{target}}\|p_{\mathrm{base}}
\right)
}.
\]
Thus, $R_{\mathrm{causal}}>0$ indicates movement toward the intended target, $R_{\mathrm{causal}}=0$ indicates no improvement, and negative values indicate movement away from the target. KL divergence is computed over the top-$100$ logits. Following the protocol used throughout the causal experiments, pairs outside the $5$th--$95$th percentile of baseline KL are excluded to limit domination by numerically unstable ratios.

Atom learning operates on standardized representations $z = (h - \mu) \oslash \sigma$, where $h$ is the raw final-token hidden state and $\mu, \sigma$ are per-dimension corpus-level statistics (see Appendix~\ref{app:implementation}). The sparse decomposition $\delta_z = \sum_{j \in \mathcal{I}_m} \alpha_j a_j$ is therefore in standardized units and is converted back to raw coordinates via
\[
\delta_h = \sigma \odot \delta_z.
\]
The modified hidden states are $h^{\mathrm{del}}_{\mathrm{SC}} = h_{\mathrm{SC}} - \delta_h$ for deletion and $h^{\mathrm{inj}}_{\mathrm{anchor}} = h_{\mathrm{anchor}} + \delta_h$ for injection, using the same $\mu$ and $\sigma$ throughout.

\paragraph{Top-$k$ KL computation.}
KL divergence is computed over the $100$ tokens with highest logit value under the target distribution $p_{\mathrm{target}}$. Both distributions are re-normalized over this shared index set before computing the divergence.

\paragraph{Inactive control construction.}
After top-$k$ encoding, atoms outside the active top-$32$ set have zero learned coefficients. The inactive control selects the $m$ atoms with largest column norms from the remaining $K - 32$ atoms and assigns each a coefficient magnitude matched to the corresponding active atom, with a random sign. The resulting vector is then rescaled to match the $\ell_2$ norm of the active intervention.

\paragraph{Percentile-based filtering.}
Pairs whose baseline $D_{\mathrm{KL}}(p_{\mathrm{anchor}} \| p_{\mathrm{SC}})$ falls outside the 5th--95th percentile range are excluded. The same mask is applied across all conditions and dose levels. Of $500$ selected pairs, approximately $450$ are retained.

\subsection{Full Causal Dose Response}
\label{app:causal_dose}

Table~\ref{tab:causal_dose_full} reports the complete deletion dose response for the global atom frame. We report medians because the normalized KL ratio can exhibit heavy-tailed behavior for a small subset of pairs. Across all seven models, active interventions strengthen as additional atoms are removed, whereas inactive and shuffled controls remain concentrated near zero. The monotonic dependence on $m$ provides evidence that semantic effects accumulate through sparse atom composition rather than being driven by a single arbitrary perturbation.

\begin{table}[H]
\centering
\small
\setlength{\tabcolsep}{3.4pt}
\renewcommand{\arraystretch}{1.08}
\begin{tabular}{llccccc}
\toprule
\textbf{Model} & \textbf{Condition}
& $m{=}1$ & $m{=}4$ & $m{=}8$ & $m{=}16$ & $m{=}32$ \\
\midrule
Mistral
& Active   & 0.013 & 0.042 & 0.071 & 0.100 & \textbf{0.125} \\
& Inactive & 0.000 & 0.001 & 0.000 & 0.002 & 0.000 \\
& Shuffled & -0.014 & -0.007 & -0.002 & -0.002 & -0.006 \\
\midrule
LLaMA
& Active   & 0.015 & 0.037 & 0.049 & 0.078 & \textbf{0.099} \\
& Inactive & 0.000 & -0.001 & -0.002 & 0.000 & -0.003 \\
& Shuffled & -0.017 & -0.003 & -0.001 & -0.005 & -0.005 \\
\midrule
Gemma
& Active   & 0.009 & 0.033 & 0.055 & 0.082 & \textbf{0.101} \\
& Inactive & 0.000 & 0.002 & 0.004 & 0.002 & 0.005 \\
& Shuffled & -0.003 & 0.005 & 0.003 & 0.005 & 0.008 \\
\midrule
Qwen
& Active   & 0.019 & 0.058 & 0.091 & 0.146 & \textbf{0.177} \\
& Inactive & 0.001 & 0.002 & 0.001 & 0.003 & 0.003 \\
& Shuffled & -0.003 & 0.002 & 0.000 & 0.001 & -0.004 \\
\midrule
GLM
& Active   & 0.005 & 0.019 & 0.037 & 0.059 & \textbf{0.074} \\
& Inactive & 0.001 & 0.002 & 0.000 & 0.003 & 0.001 \\
& Shuffled & -0.005 & -0.004 & 0.003 & 0.000 & -0.005 \\
\midrule
DeepSeek
& Active   & 0.031 & 0.100 & 0.147 & 0.216 & \textbf{0.290} \\
& Inactive & 0.000 & -0.002 & 0.002 & 0.000 & 0.002 \\
& Shuffled & -0.033 & 0.003 & -0.001 & -0.003 & -0.013 \\
\midrule
Falcon
& Active   & 0.010 & 0.055 & 0.099 & 0.119 & \textbf{0.169} \\
& Inactive & -0.010 & -0.019 & -0.009 & 0.002 & -0.007 \\
& Shuffled & -0.012 & -0.006 & -0.018 & -0.035 & -0.003 \\
\bottomrule
\end{tabular}
\caption{Full deletion dose response for the global atom frame. Values are median $R_{\mathrm{causal}}$. Active atom effects grow systematically with the number of intervened atoms, whereas norm-matched inactive atoms and coefficient-shuffled controls remain near zero.}
\label{tab:causal_dose_full}
\end{table}

The dose-response pattern is particularly clear for DeepSeek, where the active median increases from $0.031$ with one atom to $0.290$ with $32$ atoms, while the corresponding inactive and shuffled effects remain $0.002$ and $-0.013$, respectively. Similar monotonic accumulation appears for Mistral, Qwen, Falcon, LLaMA, Gemma, and GLM. Importantly, the shuffled condition uses the same learned atom directions as the active intervention but disrupts their coefficients. Its near-zero effect therefore indicates that causal behavior depends not only on selecting directions from the learned semantic frame, but also on their sparse composition for the particular semantic change.

\subsection{Global and Locally Scaled Atom Frames}
\label{app:causal_scaling}

We additionally repeat the intervention using the context-dependent atom frame
\[
A(x)=A_0\operatorname{diag}(1+r(x))
\]
introduced in Section~\ref{sec:forward}. Table~\ref{tab:causal_global_scaling} summarizes deletion and injection at $m=32$. Both parameterizations produce positive bidirectional causal effects across model families. The global frame generally yields the larger effect, while the Local frame preserves the same qualitative behavior. This is consistent with the geometric analysis in Section~\ref{sec:atom_existence}: local scaling adapts the strength of semantic axes without replacing their underlying directional organization.

\begin{table}[H]
\centering
\small
\setlength{\tabcolsep}{5pt}
\renewcommand{\arraystretch}{1.08}
\begin{tabular}{lcccc}
\toprule
& \multicolumn{2}{c}{\textbf{Global}} &
\multicolumn{2}{c}{\textbf{Local}} \\
\cmidrule(lr){2-3}\cmidrule(lr){4-5}
\textbf{Model}
& \textbf{Deletion}
& \textbf{Injection}
& \textbf{Deletion}
& \textbf{Injection} \\
\midrule
Mistral  & 0.125 & 0.105 & 0.105 & 0.084 \\
LLaMA    & 0.099 & 0.141 & 0.080 & 0.125 \\
Gemma    & 0.101 & 0.084 & 0.063 & 0.064 \\
Qwen     & 0.177 & 0.152 & 0.109 & 0.104 \\
GLM      & 0.074 & 0.084 & 0.052 & 0.054 \\
DeepSeek & 0.290 & 0.284 & 0.260 & 0.225 \\
Falcon   & 0.169 & 0.145 & 0.087 & 0.089 \\
\bottomrule
\end{tabular}
\caption{Causal effects for global and locally scaled atom frames at $m=32$. Values are median $R_{\mathrm{causal}}$ for active-atom interventions. Both frames support bidirectional intervention, while the global frame generally produces stronger effects.}
\label{tab:causal_global_scaling}
\end{table}

The persistence of causal effects under Local is important because $A(x)$ is not identical across anchors. The result indicates that context-dependent reweighting does not destroy the semantic meaning of the underlying atoms: semantic displacement reconstructed in the locally adapted frame remains capable of changing downstream model behavior in the expected direction.

\subsection{Heavy-Tailed Intervention Effects}
\label{app:causal_robust}

Because
\(
R_{\mathrm{causal}}
\)
normalizes post-intervention KL by the baseline KL of each pair, examples with small denominators can produce heavy-tailed ratios. We therefore report medians throughout the main causal comparison and additionally compute raw and winsorized means. For most models, the raw and robust statistics yield the same qualitative conclusion. Falcon3 exhibits the strongest heavy-tailed behavior: its raw mean deletion score at $m=32$ is negative despite a positive median of $0.169$, while inactive and shuffled medians remain near zero. The active median itself increases monotonically from $0.010$ to $0.169$ as $m$ grows. This discrepancy motivates the use of median $R_{\mathrm{causal}}$ as the primary cross-model statistic rather than allowing a small number of extreme normalized ratios to dominate the aggregate.

Active semantic atoms produce systematic movement toward the intended target, whereas norm-matched inactive atoms and coefficient-shuffled codes do not. Together with the bidirectional deletion/injection test and the dose-response behavior in Table~\ref{tab:causal_dose_full}, these analyses support a functional interpretation of invariant atoms as compositional directions that participate in the model's semantic computation.



\section{Additional Generalization Results}
\label{app:generalization}

This section provides complete results for the generalization experiments in Section~\ref{sec:generalization}. We test transfer to entirely unseen semantic neighborhoods, nuisance families excluded during training, and independently generated perturbation distributions.

\subsection{Generalization to Unseen Semantic Neighborhoods}
\label{app:unseen_neighborhoods}

The standard even/odd split holds out SP and SC transformations while retaining the same anchor-centered semantic neighborhoods across training and evaluation. We therefore construct a stricter group-disjoint split in which entire neighborhoods are withheld from atom learning. The $1{,}400$ groups are partitioned by a domain-stratified split into $1{,}116$ training groups and $284$ test groups. Each group contains one anchor and all of its SP/SC variants, so neither the anchor nor any associated transformation from a test group appears during training. PCA initialization is computed using training groups only. Seen performance is evaluated on held-out transformations from the training groups, whereas unseen performance uses the disjoint test groups.

This split is anchor-neighborhood-disjoint rather than topic-disjoint: semantically related topics may occur in different groups, while no test anchor or its perturbations are observed during training. The test therefore measures transfer to unseen semantic neighborhoods without claiming transfer to entirely unseen topic categories. 

The split excludes test neighborhoods from parameter fitting and PCA
initialization, while feature standardization uses the shared unlabeled
corpus statistics described in Appendix~\ref{app:implementation1}.

\begin{table}[H]
\centering
\scriptsize
\setlength{\tabcolsep}{2.8pt}
\renewcommand{\arraystretch}{1.05}
\begin{tabular}{llcccccc}
\toprule
& &
\multicolumn{2}{c}{$E_{\mathrm{rec}}\downarrow$} &
\multicolumn{2}{c}{$\mathrm{Cos}\uparrow$} &
\multicolumn{2}{c}{$S/T\uparrow$} \\
\cmidrule(lr){3-4}\cmidrule(lr){5-6}\cmidrule(lr){7-8}
\textbf{Model}
& \textbf{Frame}
& \textbf{Seen}
& \textbf{Unseen}
& \textbf{Seen}
& \textbf{Unseen}
& \textbf{Seen}
& \textbf{Unseen} \\
\midrule
Mistral
& Global  & 0.709 & 0.723 & 0.695 & 0.680 & 1.29 & 1.17 \\
& Local & 0.695 & 0.714 & 0.708 & 0.689 & 1.43 & 1.10 \\
\midrule
LLaMA
& Global  & 0.733 & 0.747 & 0.665 & 0.650 & 1.85 & 1.68 \\
& Local & 0.722 & 0.740 & 0.676 & 0.658 & 2.01 & 1.54 \\
\midrule
Gemma
& Global  & 0.718 & 0.732 & 0.679 & 0.665 & 1.67 & 1.30 \\
& Local & 0.699 & 0.717 & 0.697 & 0.680 & 2.28 & 1.13 \\
\midrule
Qwen
& Global  & 0.688 & 0.703 & 0.714 & 0.700 & 1.62 & 1.41 \\
& Local & 0.671 & 0.690 & 0.729 & 0.712 & 1.85 & 1.36 \\
\midrule
GLM
& Global  & 0.722 & 0.739 & 0.676 & 0.658 & 2.00 & 1.76 \\
& Local & 0.711 & 0.732 & 0.687 & 0.666 & 2.18 & 1.65 \\
\midrule
DeepSeek
& Global  & 0.699 & 0.712 & 0.704 & 0.690 & 1.06 & 0.96 \\
& Local & 0.685 & 0.702 & 0.716 & 0.700 & 1.32 & 0.94 \\
\midrule
Falcon
& Global  & 0.770 & 0.795 & 0.621 & 0.592 & 1.69 & 1.56 \\
& Local & 0.761 & 0.790 & 0.632 & 0.598 & 1.77 & 1.49 \\
\bottomrule
\end{tabular}
\caption{Generalization to group-disjoint semantic neighborhoods. Entire anchor-centered groups are withheld from atom learning. Reconstruction remains stable on unseen groups, while semantic selectivity transfers strongly for most models.}
\label{tab:unseen_neighborhood_full}
\end{table}

Reconstruction transfers consistently across architectures. Unseen-group cosine retains approximately $95$--$98\%$ of seen-group performance, with only modest increases in $L_{\mathrm{sem}}$. The semantic selectivity ratio also remains above one for six of seven models under both Global and Local. DeepSeek is the boundary case, decreasing from $1.06$ to $0.96$ for Global and from $1.32$ to $0.94$ for Local. Thus, reusable semantic reconstruction is highly stable across unseen neighborhoods, while semantic--nuisance separation is somewhat more sensitive to content shift.

Global also tends to retain a larger fraction of its seen-group $S/T$ than Local. This is consistent with the distinction developed in Appendix~\ref{app:global_local}: the shared frame is directly reusable across inputs, whereas the locally conditioned model introduces additional anchor dependence. Importantly, both parameterizations retain nearly all of their reconstruction cosine on unseen groups.

\subsection{Held-Out Paraphrase Families}
\label{app:heldout_styles}

We next evaluate nuisance-family generalization by excluding entire SP transformation families during training. Split A holds out question and passive transformations, Split B holds out casual and technical transformations, and Split C holds out simplified and verbose transformations. Each model is retrained using the remaining six SP families, while SC examples retain the standard train/test construction.

We define
\[
R_{\mathrm{transfer}}
=
\frac{(S/T)_{\mathrm{held\text{-}out}}}
     {(S/T)_{\mathrm{train}}},
\]
where values near one indicate preservation of semantic selectivity under nuisance families absent during optimization.

\begin{table}[h]
\centering
\scriptsize
\setlength{\tabcolsep}{3.0pt}
\renewcommand{\arraystretch}{1.05}
\begin{tabular}{llcccccc}
\toprule
& & \multicolumn{3}{c}{\textbf{Global}}
& \multicolumn{3}{c}{\textbf{Local}} \\
\cmidrule(lr){3-5}\cmidrule(lr){6-8}
\textbf{Model}
& \textbf{Split}
& \textbf{Train}
& \textbf{Held}
& \textbf{Transfer}
& \textbf{Train}
& \textbf{Held}
& \textbf{Transfer} \\
\midrule
Mistral
& A & 1.86 & 2.42 & 1.30 & 2.52 & 3.60 & 1.43 \\
& B & 2.01 & 1.63 & 0.81 & 2.76 & 2.23 & 0.81 \\
& C & 2.07 & 1.29 & 0.62 & 2.87 & 1.78 & 0.62 \\
\midrule
LLaMA
& A & 2.63 & 3.32 & 1.26 & 3.40 & 4.78 & 1.41 \\
& B & 2.80 & 2.37 & 0.85 & 3.67 & 3.11 & 0.85 \\
& C & 3.01 & 1.88 & 0.62 & 3.95 & 2.45 & 0.62 \\
\midrule
Gemma
& A & 2.29 & 2.49 & 1.09 & 4.83 & 5.79 & 1.20 \\
& B & 2.36 & 2.12 & 0.90 & 5.18 & 4.71 & 0.91 \\
& C & 2.53 & 1.71 & 0.67 & 5.39 & 3.71 & 0.69 \\
\midrule
Qwen
& A & 2.54 & 2.70 & 1.06 & 3.38 & 3.83 & 1.13 \\
& B & 2.47 & 2.08 & 0.84 & 2.73 & 2.34 & 0.86 \\
& C & 2.63 & 1.60 & 0.61 & 3.46 & 2.11 & 0.61 \\
\midrule
GLM
& A & 2.77 & 3.31 & 1.19 & 3.73 & 4.78 & 1.28 \\
& B & 2.99 & 2.56 & 0.86 & 3.90 & 3.40 & 0.87 \\
& C & 3.22 & 1.93 & 0.60 & 4.20 & 2.53 & 0.60 \\
\midrule
DeepSeek
& A & 1.57 & 1.86 & 1.19 & 2.39 & 3.04 & 1.27 \\
& B & 1.58 & 1.47 & 0.93 & 2.45 & 2.25 & 0.92 \\
& C & 1.69 & 1.13 & 0.67 & 2.64 & 1.75 & 0.66 \\
\midrule
Falcon
& A & 2.14 & 2.56 & 1.20 & 2.75 & 3.57 & 1.30 \\
& B & 2.16 & 1.92 & 0.89 & 2.84 & 2.51 & 0.88 \\
& C & 2.28 & 1.55 & 0.68 & 2.98 & 2.00 & 0.67 \\
\bottomrule
\end{tabular}
\caption{Generalization to SP families excluded during training. Entries report $S/T$ on observed and held-out nuisance families and their ratio. Split A holds out question/passive, Split B casual/technical, and Split C simplified/verbose.}
\label{tab:style_generalization_full}
\end{table}

The pattern is highly consistent across architectures. Split A transfers particularly well and often yields $R_{\mathrm{transfer}}>1$, indicating that question and passive transformations are suppressed at least as strongly as observed nuisance families. Split B exhibits moderate degradation, whereas Split C is consistently the most difficult, with transfer ratios around $0.60$--$0.69$. Nevertheless, held-out $S/T$ remains above one in every model, split, and frame variant. Across all conditions, held-out $S/T$ ranges from $1.13$ to $3.32$ for Global and from $1.75$ to $5.79$ for Local. Thus, invariance generalizes to nuisance mechanisms not used during learning, although the degree of preservation depends on the transformation family.

\subsection{Bidirectional Cross-Generator Transfer}
\label{app:cross_generator}

Finally, we test whether the learned semantic geometry depends on the model used to generate the controlled perturbations. The primary dataset is generated by Mistral-7B-Instruct-v0.3, while an independent perturbation set is generated by Qwen2.5-7B-Instruct. In each experiment, both perturbation sets are encoded by the same representation model; only the perturbation generator changes. Training-set normalization statistics are reused at evaluation so that both sets remain in a common representation coordinate system.

We evaluate both directions. In Mistral$\rightarrow$Qwen, the atom frame is learned from Mistral-generated perturbations and evaluated on Qwen-generated perturbations. In Qwen$\rightarrow$Mistral, the roles are reversed. PCA initialization is recomputed from the corresponding training generator in each direction.

Mistral$\rightarrow$Qwen transfer is uniformly strong. Unseen-generator $S/T$ exceeds its seen-generator value for every representation model, with transfer ratios of $1.11$--$1.33$ for Global and $1.09$--$1.75$ for Local. The reverse direction is systematically weaker: Qwen$\rightarrow$Mistral transfer ranges from $0.31$ to $0.68$ for Global and from $0.36$ to $0.71$ for Local. Importantly, however, unseen-generator $S/T$ remains above one in every model and direction.

The asymmetry shows that generator transfer is not distribution-free. Frames learned from Qwen-generated perturbations retain semantic selectivity on Mistral-generated perturbations but lose a substantial fraction of their training-domain ratio, whereas Mistral-trained frames transfer without such degradation. We therefore interpret this experiment as evidence that the learned semantic coordinates survive perturbation-source shift, while their quantitative selectivity remains sensitive to the statistics of the generator-specific perturbation distribution.

\begin{table}[H]
\centering
\scriptsize
\setlength{\tabcolsep}{2.8pt}
\renewcommand{\arraystretch}{1.05}
\begin{tabular}{llcccc}
\toprule
\textbf{Model}
& \textbf{Direction}
& \textbf{Frame}
& \textbf{Seen $S/T$}
& \textbf{Unseen $S/T$}
& \textbf{Transfer} \\
\midrule
Mistral
& M$\rightarrow$Q & Global  & 1.23 & 1.46 & 1.19 \\
&                 & Local & 1.62 & 1.87 & 1.15 \\
& Q$\rightarrow$M & Global  & 4.09 & 2.02 & 0.49 \\
&                 & Local & 7.60 & 4.48 & 0.59 \\
\midrule
LLaMA
& M$\rightarrow$Q & Global  & 1.74 & 2.10 & 1.21 \\
&                 & Local & 2.10 & 2.47 & 1.17 \\
& Q$\rightarrow$M & Global  & 5.06 & 2.33 & 0.46 \\
&                 & Local & 10.37 & 5.63 & 0.54 \\
\midrule
Gemma
& M$\rightarrow$Q & Global  & 1.42 & 1.88 & 1.33 \\
&                 & Local & 3.00 & 3.66 & 1.22 \\
& Q$\rightarrow$M & Global  & 8.94 & 2.79 & 0.31 \\
&                 & Local & 20.79 & 7.55 & 0.36 \\
\midrule
Qwen
& M$\rightarrow$Q & Global  & 1.48 & 1.86 & 1.26 \\
&                 & Local & 1.28 & 2.23 & 1.75 \\
& Q$\rightarrow$M & Global  & 4.88 & 2.18 & 0.45 \\
&                 & Local & 9.73 & 5.15 & 0.53 \\
\midrule
GLM
& M$\rightarrow$Q & Global  & 1.90 & 2.36 & 1.24 \\
&                 & Local & 2.32 & 2.79 & 1.20 \\
& Q$\rightarrow$M & Global  & 5.44 & 2.44 & 0.45 \\
&                 & Local & 10.44 & 5.40 & 0.52 \\
\midrule
DeepSeek
& M$\rightarrow$Q & Global  & 1.04 & 1.22 & 1.18 \\
&                 & Local & 1.53 & 1.71 & 1.12 \\
& Q$\rightarrow$M & Global  & 3.85 & 1.84 & 0.48 \\
&                 & Local & 6.91 & 4.15 & 0.60 \\
\midrule
Falcon
& M$\rightarrow$Q & Global  & 1.62 & 1.80 & 1.11 \\
&                 & Local & 1.88 & 2.05 & 1.09 \\
& Q$\rightarrow$M & Global  & 2.95 & 2.00 & 0.68 \\
&                 & Local & 5.46 & 3.90 & 0.71 \\
\bottomrule
\end{tabular}
\caption{Bidirectional cross-generator generalization. $M$ and $Q$ denote Mistral- and Qwen-generated perturbation sets. Transfer is the ratio between unseen-generator and seen-generator $S/T$.}
\label{tab:cross_generator_full}
\end{table}



\section{Additional Retrieval Results}
\label{app:retrieval}

This section expands the retrieval analysis in Section~\ref{sec:functional_validation}. We report the complete comparison among raw representations, PCA, Global signatures, Local signatures, consistency-regularized variants, whitening and normalization corrections, and an oracle shared-local-frame construction. We additionally analyze retrieval by SP style and the coordinate mismatch introduced by input-dependent local scaling.

\subsection{Full Retrieval Comparison}
\label{app:retrieval_full}

The retrieval gallery contains the $1{,}400$ anchor representations, while the held-out odd-indexed SP variants serve as queries. A query is considered correct when the retrieved anchor belongs to the same semantic group. All projected signatures are $256$ dimensional, whereas Raw uses the original model hidden dimension.

For the shared global atom frame,
\[
s_{\mathrm{global}}(x)=A_0^\top z(x).
\]
For the locally scaled frame,
\[
s_{\mathrm{local}}(x)=A(x)^\top z(x),
\qquad
A(x)=A_0\operatorname{diag}(1+r(x)).
\]
We additionally evaluate two consistency-regularized Local variants, post-hoc coordinate standardization, ScaleNorm, and an Oracle construction that uses the anchor's local frame for both anchor and query.

We additionally penalize differences between the modulation
vectors of semantically equivalent inputs:
\[
\mathcal{L}_{\mathrm{cons}}
=
\frac{1}{M}\sum_{m=1}^{M}
\left\|r(x)-r(x_m^{\mathrm{SP}})\right\|_2^2.
\]
Here $r(x)=\tanh(g_\eta(z))$, as defined in Section 3.2.
Local+$\mathcal{L}_{\mathrm{cons}}(\lambda)$ denotes
the corresponding variant with regularization weight
$\lambda$.

\begin{table}[H]
\centering
\scriptsize
\setlength{\tabcolsep}{2.6pt}
\renewcommand{\arraystretch}{1.04}
\begin{tabular}{llcccccccc}
\toprule
& & \multicolumn{2}{c}{\textbf{Mistral}}
& \multicolumn{2}{c}{\textbf{LLaMA}}
& \multicolumn{2}{c}{\textbf{Gemma}}
& \multicolumn{2}{c}{\textbf{Qwen}} \\
\cmidrule(lr){3-4}\cmidrule(lr){5-6}\cmidrule(lr){7-8}\cmidrule(lr){9-10}
\textbf{Method} & \textbf{Dim.}
& \textbf{R@1} & \textbf{mAP}
& \textbf{R@1} & \textbf{mAP}
& \textbf{R@1} & \textbf{mAP}
& \textbf{R@1} & \textbf{mAP} \\
\midrule
Raw
& $d$
& .440 & .519
& .419 & .508
& .384 & .472
& .287 & .363 \\
PCA
& 256
& .292 & .366
& .294 & .381
& .242 & .329
& .204 & .273 \\
Global
& 256
& \textbf{.458} & \textbf{.552}
& \textbf{.403} & \textbf{.510}
& \textbf{.345} & \textbf{.447}
& .291 & .388 \\
Local
& 256
& .370 & .462
& .309 & .413
& .230 & .324
& .228 & .319 \\
Local+$L_{\mathrm{cons}}(.05)$
& 256
& .375 & .468
& .313 & .415
& .231 & .325
& .228 & .319 \\
Local+$L_{\mathrm{cons}}(.10)$
& 256
& .375 & .470
& .312 & .416
& .231 & .325
& .230 & .321 \\
Local-Whiten
& 256
& .412 & .502
& .340 & .438
& .241 & .333
& \textbf{.324} & \textbf{.418} \\
Local-ScaleNorm
& 256
& .326 & .413
& .309 & .409
& .244 & .339
& .222 & .310 \\
Oracle
& 256
& .416 & .510
& .358 & .464
& .262 & .359
& .260 & .352 \\
\bottomrule
\end{tabular}
\caption{Retrieval results for Mistral, LLaMA, Gemma, and Qwen. Bold marks the strongest $256$-dimensional method for each model. Raw uses the native hidden dimension $d$; all other methods use $256$ coordinates.}
\label{tab:retrieval_full_a}
\end{table}

\begin{table}[H]
\centering
\scriptsize
\setlength{\tabcolsep}{2.8pt}
\renewcommand{\arraystretch}{1.04}
\begin{tabular}{llcccccc}
\toprule
& & \multicolumn{2}{c}{\textbf{GLM}}
& \multicolumn{2}{c}{\textbf{DeepSeek}}
& \multicolumn{2}{c}{\textbf{Falcon}} \\
\cmidrule(lr){3-4}\cmidrule(lr){5-6}\cmidrule(lr){7-8}
\textbf{Method} & \textbf{Dim.}
& \textbf{R@1} & \textbf{mAP}
& \textbf{R@1} & \textbf{mAP}
& \textbf{R@1} & \textbf{mAP} \\
\midrule
Raw
& $d$
& .411 & .505
& .407 & .491
& .403 & .498 \\
PCA
& 256
& .286 & .375
& .248 & .330
& .258 & .344 \\
Global
& 256
& \textbf{.363} & \textbf{.464}
& \textbf{.368} & \textbf{.466}
& \textbf{.314} & \textbf{.414} \\
Local
& 256
& .288 & .386
& .260 & .353
& .241 & .333 \\
Local+$L_{\mathrm{cons}}(.05)$
& 256
& .291 & .389
& .261 & .355
& .241 & .332 \\
Local+$L_{\mathrm{cons}}(.10)$
& 256
& .292 & .390
& .261 & .356
& .241 & .333 \\
Local-Whiten
& 256
& .335 & .430
& .317 & .408
& .244 & .336 \\
Local-ScaleNorm
& 256
& .279 & .376
& .273 & .368
& .247 & .338 \\
Oracle
& 256
& .326 & .428
& .294 & .391
& .279 & .374 \\
\bottomrule
\end{tabular}
\caption{Retrieval results for GLM, DeepSeek, and Falcon. Global is the strongest unmodified $256$-dimensional signature on all three models, substantially outperforming PCA at the same dimensionality.}
\label{tab:retrieval_full_b}
\end{table}

The complete results reinforce the main-paper conclusion. Global consistently dominates PCA at the same dimensionality and is the strongest directly comparable atom signature across most model families. Mistral is the clearest case, where the $256$-dimensional Global signature exceeds the original $4096$-dimensional representation on both R@1 and mAP. LLaMA retains nearly identical R@1 and slightly improves mAP, while Qwen also improves over Raw under the shared frame. In the remaining models, compression incurs a moderate retrieval cost but preserves substantially more semantic identity than PCA.

The locally scaled signatures are consistently weaker than Global under direct nearest-neighbor comparison. The consistency regularizer produces only small gains, indicating that encouraging similar scaling vectors for anchor and paraphrase is insufficient to fully align their local coordinate systems. Whitening provides a larger recovery for several models and is especially effective for Qwen, where R@1 rises from $0.228$ under Local to $0.324$, exceeding both Global and Raw.

\subsection{Recall by Semantic-Preserving Style}
\label{app:retrieval_styles}

To determine whether retrieval behavior is dominated by particular nuisance transformations, we report R@1 separately for each SP style. Table~\ref{tab:retrieval_style_a} shows representative results for Mistral, LLaMA, Qwen, and GLM.

The style-wise results clarify when invariant coordinates are most useful. Raw representations are often strongest for relatively mild transformations such as formal or question reformulations. In contrast, Global frequently closes or reverses the gap on styles that induce larger nuisance motion. Mistral improves from $0.305$ to $0.414$ on imperative queries and from $0.293$ to $0.330$ on verbose queries. LLaMA improves on passive and imperative transformations, while Qwen improves substantially on passive and imperative styles. GLM similarly improves on passive transformations. This pattern is consistent with the geometric objective: suppressing nuisance variation becomes most useful when the surface transformation causes a relatively large displacement in the original representation space.

\begin{table}[H]
\centering
\scriptsize
\setlength{\tabcolsep}{2.6pt}
\renewcommand{\arraystretch}{1.04}
\begin{tabular}{lcccccccc}
\toprule
& \multicolumn{2}{c}{\textbf{Mistral}}
& \multicolumn{2}{c}{\textbf{LLaMA}}
& \multicolumn{2}{c}{\textbf{Qwen}}
& \multicolumn{2}{c}{\textbf{GLM}} \\
\cmidrule(lr){2-3}\cmidrule(lr){4-5}\cmidrule(lr){6-7}\cmidrule(lr){8-9}
\textbf{Style}
& \textbf{Raw} & \textbf{Global}
& \textbf{Raw} & \textbf{Global}
& \textbf{Raw} & \textbf{Global}
& \textbf{Raw} & \textbf{Global} \\
\midrule
Formal     & .698 & .681 & .699 & .611 & .578 & .505 & .663 & .555 \\
Casual     & .430 & .422 & .460 & .422 & .219 & .228 & .445 & .366 \\
Question   & .680 & .671 & .654 & .629 & .552 & .500 & .679 & .611 \\
Technical  & .423 & .433 & .403 & .360 & .333 & .292 & .398 & .329 \\
Simplified & .486 & .469 & .440 & .369 & .254 & .223 & .400 & .307 \\
Verbose    & .293 & .330 & .263 & .259 & .169 & .188 & .259 & .231 \\
Passive    & .472 & .450 & .352 & .386 & .109 & .190 & .176 & .248 \\
Imperative & .305 & .414 & .300 & .365 & .256 & .334 & .411 & .391 \\
\bottomrule
\end{tabular}
\caption{R@1 by SP style for representative models. Global often provides its largest relative gains on harder nuisance transformations such as verbose, passive, and imperative reformulations.}
\label{tab:retrieval_style_a}
\end{table}

Local remains below Global across essentially all styles. This supports the interpretation that the retrieval gap is not caused by a single nuisance family but by the cross-example coordinate mismatch introduced by input-dependent local frames.

\subsection{Diagnosing the Global--Local Signature Gap}
\label{app:retrieval_mismatch}

The difference between Global and Local can be understood directly from their coordinate systems. For two semantically equivalent inputs $x$ and $x'$, the shared signature compares
\[
A_0^\top z(x)
\qquad\text{and}\qquad
A_0^\top z(x'),
\]
so both vectors are expressed in the same coordinates. Under local scaling, however,
\[
A(x)^\top z(x)
\qquad\text{and}\qquad
A(x')^\top z(x'),
\]
are compared even though
\[
A(x)
=
A_0\operatorname{diag}(1+r(x))
\neq
A_0\operatorname{diag}(1+r(x'))
=
A(x').
\]
Thus, semantically equivalent examples may differ not only because their hidden states differ, but also because the coordinate-wise scales used to represent them differ.

The Oracle construction isolates this effect by using the anchor frame for both sides,
\[
s_{\mathrm{oracle}}(x')
=
A(x_{\mathrm{anchor}})^\top z(x').
\]
Oracle consistently improves over ordinary Local, confirming that sharing the local frame reduces the retrieval penalty. For example, Mistral R@1 increases from $0.370$ to $0.416$, LLaMA from $0.309$ to $0.358$, and Falcon from $0.241$ to $0.279$. The recovery is partial rather than complete, indicating that local scaling affects both coordinate comparability and the geometry of the projected representation itself.

Post-hoc whitening provides a complementary diagnostic. Let
\[
\tilde{s}(x)
=
\frac{s_{\mathrm{local}}(x)-\mu}{\sigma},
\]
where $\mu$ and $\sigma$ are estimated from anchor signatures. Whitening improves Local in most model families, suggesting that part of the mismatch arises from coordinate-wise mean and variance shifts induced by input-dependent reweighting. The effect is particularly strong for Qwen, where Local-Whiten reaches R@1 $0.324$, compared with $0.228$ for Local and $0.291$ for Global. By contrast, 
dividing out the predicted scaling factors through ScaleNorm removes the explicit diagonal modulation but does not recover Stage-1 Global retrieval performance. This can arise because the Stage-2 shared frame $A_0^{(\text{Local})}$ co-adapts with the modulation network and may differ from the Stage-1 frame $A_0^{(\text{Global})}$, so inverse rescaling alone does not recover the original coordinate system.

Together, these diagnostics sharpen the distinction between the two representations. The global frame provides a common coordinate system and is therefore naturally suited to cross-example similarity. The local frame improves semantic decomposition within a neighborhood, but its context-dependent reweighting reduces direct comparability across different inputs. This is consistent with the global-to-local geometry established in Section~\ref{sec:training}: local modulation refines how shared semantic directions are expressed without replacing the value of a globally shared coordinate system for comparison tasks.



\section{Atom Signatures under Model Modification}
\label{app:model_modification}

This section examines whether atom-based representations remain semantically meaningful after common model modifications. We consider three variants of each base language model: the unmodified model, a LoRA fine-tuned model, and a distilled LoRA variant. The purpose is not to develop a standalone model-attribution method, but to test whether the learned semantic coordinates remain stable while retaining sensitivity to model-specific changes.

For fine-tuning, we apply LoRA adaptation on a $10$K subset of Alpaca. The distilled variant uses the same LoRA parameterization but is trained with a combined supervised and teacher-matching objective. We compare six representations: the raw hidden state, PCA, a random orthogonal projection, the shared Global atom signature, the locally scaled signature, and a sparsified TopK atom signature retaining the top $32$ coordinates.

\subsection{Signature Drift under Model Modification}
\label{app:model_mod_drift}

Semantic preservation does not imply that the signatures are numerically identical across model variants. We therefore measure pairwise drift between corresponding signatures using cosine similarity and Euclidean distance. Table~\ref{tab:model_mod_drift} reports the Global and Local representations together with Raw for reference.

\begin{table}[H]
\centering
\scriptsize
\setlength{\tabcolsep}{2.8pt}
\renewcommand{\arraystretch}{1.05}
\begin{tabular}{llccccc}
\toprule
\textbf{Model}
& \textbf{Signature}
& \multicolumn{2}{c}{\textbf{Base$\rightarrow$FT}}
& \multicolumn{2}{c}{\textbf{Base$\rightarrow$Distill}} \\
\cmidrule(lr){3-4}\cmidrule(lr){5-6}
& & $\cos\uparrow$ & $L_2\downarrow$
& $\cos\uparrow$ & $L_2\downarrow$ \\
\midrule
Mistral
& Raw     & .770 & 41.6 & .932 & 22.5 \\
& Global  & .811 & 1.89 & .950 & .97 \\
& Local & .817 & .99 & .954 & .50 \\
\midrule
LLaMA
& Raw     & .906 & 26.8 & .953 & 18.9 \\
& Global  & .942 & 1.11 & .976 & .72 \\
& Local & .950 & .71 & .982 & .43 \\
\midrule
Gemma
& Raw     & .964 & 14.8 & .983 & 10.2 \\
& Global  & .977 & .62 & .990 & .42 \\
& Local & .983 & .33 & .995 & .19 \\
\midrule
Qwen
& Raw     & .938 & 18.9 & .953 & 16.7 \\
& Global  & .960 & .85 & .972 & .72 \\
& Local & .967 & .43 & .978 & .35 \\
\midrule
GLM
& Raw     & .938 & 21.7 & .963 & 16.8 \\
& Global  & .966 & .84 & .983 & .61 \\
& Local & .970 & .85 & .987 & .57 \\
\midrule
DeepSeek
& Raw     & .872 & 22.0 & .929 & 16.2 \\
& Global  & .889 & 1.25 & .943 & .89 \\
& Local & .898 & .64 & .951 & .45 \\
\midrule
Falcon
& Raw     & .928 & 20.2 & .944 & 17.9 \\
& Global  & .954 & .92 & .972 & .79 \\
& Local & .959 & .79 & .972 & .67 \\
\bottomrule
\end{tabular}
\caption{Signature drift under fine-tuning and distillation. Atom signatures exhibit high cosine stability and substantially smaller Euclidean drift than raw hidden representations.}
\label{tab:model_mod_drift}
\end{table}

Two patterns are consistent across architectures. First, distillation generally preserves signatures more strongly than fine-tuning. Second, Global and Local signatures exhibit much smaller Euclidean drift than the original hidden states. For example, Mistral base-to-fine-tuned drift decreases from $41.6$ in the raw representation to $1.89$ under Global and $0.99$ under Local. LLaMA decreases from $26.8$ to $1.11$ and $0.71$, while Gemma decreases from $14.8$ to $0.62$ and $0.33$.

These differences should not be interpreted as direct cross-space comparisons of absolute scale, since the raw and projected representations have different dimensions and coordinate magnitudes. The more informative observation is the consistently high cosine correspondence together with semantic stability from Section~\ref{app:model_mod_semantic}. The learned atom coordinates therefore remain well aligned after modification even though the underlying model parameters have changed.

Local typically exhibits the smallest drift. This is consistent with its role as a locally adaptive representation: context-dependent reweighting can absorb part of the representational change induced by fine-tuning or distillation. This stability is complementary to the retrieval result in Appendix~\ref{app:retrieval}, where Global is preferable for direct cross-example comparison because it provides a single shared coordinate system.

\subsection{Semantic Stability under Fine-Tuning and Distillation}
\label{app:model_mod_semantic}

We first ask whether semantic structure encoded by the signatures survives model modification.
\begin{table}[H]
\centering
\scriptsize
\setlength{\tabcolsep}{3.0pt}
\renewcommand{\arraystretch}{1.05}
\begin{tabular}{llcccc}
\toprule
\textbf{Model}
& \textbf{Signature}
& \textbf{Dim.}
& \textbf{Base}
& \textbf{Fine-tuned}
& \textbf{Distilled} \\
\midrule
Mistral
& Raw     & 4096 & .878 & .882 & .887 \\
& PCA     & 256  & .849 & .852 & .855 \\
& Global  & 256  & .849 & .852 & .859 \\
& Local & 256  & .833 & .843 & .843 \\
& TopK    & 256  & .809 & .822 & .823 \\
\midrule
LLaMA
& Raw     & 4096 & .892 & .893 & .891 \\
& PCA     & 256  & .876 & .876 & .874 \\
& Global  & 256  & .872 & .873 & .870 \\
& Local & 256  & .867 & .868 & .865 \\
& TopK    & 256  & .839 & .844 & .840 \\
\midrule
Qwen
& Raw     & 3584 & .879 & .882 & .882 \\
& PCA     & 256  & .866 & .864 & .863 \\
& Global  & 256  & .866 & .861 & .861 \\
& Local & 256  & .853 & .855 & .852 \\
& TopK    & 256  & .832 & .832 & .834 \\
\midrule
GLM
& Raw     & 4096 & .887 & .886 & .885 \\
& PCA     & 256  & .870 & .869 & .870 \\
& Global  & 256  & .869 & .866 & .866 \\
& Local & 256  & .862 & .858 & .859 \\
& TopK    & 256  & .838 & .834 & .833 \\
\midrule
DeepSeek
& Raw     & 2048 & .876 & .877 & .877 \\
& PCA     & 256  & .854 & .850 & .853 \\
& Global  & 256  & .851 & .851 & .851 \\
& Local & 256  & .842 & .842 & .841 \\
& TopK    & 256  & .810 & .814 & .816 \\
\midrule
Falcon
& Raw     & 3072 & .880 & .851 & .874 \\
& PCA     & 256  & .861 & .851 & .851 \\
& Global  & 256  & .857 & .850 & .853 \\
& Local & 256  & .851 & .846 & .845 \\
& TopK    & 256  & .818 & .812 & .812 \\
\midrule
Gemma
& Raw     & 3584 & .892 & .894 & .892 \\
& PCA     & 256  & .879 & .876 & .877 \\
& Global  & 256  & .868 & .866 & .866 \\
& Local & 256  & .863 & .864 & .865 \\
& TopK    & 256  & .843 & .842 & .843 \\
\bottomrule
\end{tabular}
\caption{Zero-shot SC-type classification after model modification. The classifier is trained only on base-model signatures and evaluated without retraining on fine-tuned and distilled variants. Atom-based signatures retain nearly unchanged semantic classification performance across modifications.}
\label{tab:model_mod_semantic}
\end{table}

 A seven-way SC-type classifier is trained using signatures from the base model and then evaluated \emph{zero-shot}, without retraining, on corresponding signatures from the fine-tuned and distilled variants. Thus, performance preservation indicates that the semantic organization of the representation remains aligned across model modifications.

Across architectures, semantic classification changes only modestly after fine-tuning or distillation. For example, Global changes from $0.849$ to $0.852$ and $0.859$ on Mistral, from $0.872$ to $0.873$ and $0.870$ on LLaMA, and from $0.851$ to $0.851$ and $0.851$ on DeepSeek. Even Falcon, which exhibits the largest degradation among the tested models, changes only from $0.857$ to $0.850$ and $0.853$. These results indicate that the semantic organization captured by the atom coordinates is largely preserved under moderate parameter adaptation.

Importantly, the Global representation remains close to PCA and Raw while using only $256$ dimensions. Thus, the stability is not merely a property of the full hidden representation; the compact atom coordinates themselves preserve the semantic structure required for SC discrimination.

\subsection{Model-Source Classification}
\label{app:model_mod_source}

We finally ask whether semantic stability eliminates all information about the underlying model variant. We train a three-way classifier to distinguish signatures from the base, fine-tuned, and distilled models. Equal-sized signature pools from the three variants are randomly divided into $80/20$ train/validation splits.

Model-source information remains readily detectable in the dense signatures. Global reaches $0.986$ accuracy on Mistral, $0.942$ on LLaMA, $0.893$ on Qwen, $0.916$ on GLM, and $0.965$ on DeepSeek. However, this behavior is not unique to invariant atoms: PCA and even random projections also achieve high source-classification accuracy. We therefore interpret this experiment conservatively. The result does not establish a specialized fingerprinting advantage of the atom representation; rather, it shows that preserving semantic structure does not erase model-specific variation.

\begin{table}[H]
\centering
\scriptsize
\setlength{\tabcolsep}{3.2pt}
\renewcommand{\arraystretch}{1.05}
\begin{tabular}{lcccccc}
\toprule
\textbf{Model}
& \textbf{Raw}
& \textbf{PCA}
& \textbf{Random}
& \textbf{Global}
& \textbf{Local}
& \textbf{TopK} \\
\midrule
Mistral  & 1.000 & .986 & .984 & .986 & .974 & .862 \\
LLaMA    & .999 & .937 & .940 & .942 & .907 & .684 \\
Gemma    & .992 & .879    & .914   & .898    & .798 & .587 \\
Qwen     & .993 & .887 & .891 & .893 & .852 & .636 \\
GLM      & .998 & .909 & .916 & .916 & .866 & .632 \\
DeepSeek & .998 & .958 & .958 & .965 & .934 & .769 \\
Falcon   & .986 & .810 & .805 & .806 & .768 & .591 \\
\bottomrule
\end{tabular}
\caption{Three-way model-source classification accuracy for base, fine-tuned, and distilled variants. Dense $256$-dimensional atom signatures retain substantial model-specific information, whereas aggressive TopK sparsification removes a larger fraction of that signal.}
\label{tab:model_mod_source}
\end{table}

TopK signatures exhibit substantially lower model-source accuracy than their dense counterparts. For example, accuracy decreases from $0.942$ to $0.684$ on LLaMA, from $0.893$ to $0.636$ on Qwen, and from $0.916$ to $0.632$ on GLM. This suggests that model-specific variation is distributed partly through lower-magnitude coordinates that are discarded by aggressive sparsification, whereas the dominant sparse coordinates retain more of the shared semantic structure.

Taken together, these experiments reveal a useful separation. Atom signatures remain semantically stable under fine-tuning and distillation, while dense signatures still contain sufficient residual variation to distinguish modified model variants. We view this as a robustness property of the representation rather than as a primary model-attribution contribution.


\section{Boundary Case Evaluation on External Perturbations}
\label{sec:boundary}

Surface-level edit size does not determine semantic status: small
changes in wording, syntax, or punctuation may preserve meaning or
alter it substantially. Our SP/SC distinction concerns the resulting
meaning in context, rather than the linguistic form of the edit.
To test whether the learned atom frames remain selective when surface
changes are similar, we evaluate on an external corpus containing
minimal semantic changes and closely matched paraphrases.

\paragraph{Dataset.}
The corpus contains 210 perturbation pairs across 21 anchor queries,
organized into three subsets.
The ``simple\_mixed'' subset contains 10 pairs derived from a
single sentence, with minimal lexical edits for both paraphrases and
semantic changes.
The ``pilot\_softSC'' and ``pilot\_mixed'' subsets each
contain 100 pairs spanning 10 anchor queries across diverse domains.
The semantic changes use five strategies:
``entity\_swap'', ``predicate\_swap'',
``phenomenon\_swap'', ``process\_swap'', and
``topic\_swap''.
Each semantic change alters only a single word or short phrase;
for example, ``Describe the \textit{hunting} behavior of domestic
cats'' becomes ``Describe the \textit{sleeping} behavior of domestic
cats.''
These edits make lexical similarity alone an unreliable indicator
of meaning preservation.
The external perturbations are excluded from training, and their
representations are standardized using the training-data statistics.

\paragraph{Results.}
Table~\ref{tab:boundary} reports reconstruction cosine similarity,
semantic selectivity ($S/T$), and nuisance energy for Qwen2.5-7B
and Mistral-7B, using the evaluation definitions introduced earlier.
We retain the Global and Local terminology used throughout the paper;
Local denotes the model with anchor-dependent diagonal modulation.

\begin{table}[!htbp]
\centering
\scriptsize
\setlength{\tabcolsep}{3.3pt}
\renewcommand{\arraystretch}{1.05}
\begin{tabular}{lcccccc}
\toprule
& \multicolumn{2}{c}{\textbf{Cos} $\uparrow$}
& \multicolumn{2}{c}{$S/T$ $\uparrow$}
& \multicolumn{2}{c}{\textbf{Nui.\ Energy} $\downarrow$} \\
\cmidrule(lr){2-3}
\cmidrule(lr){4-5}
\cmidrule(lr){6-7}
\textbf{Method}
& \textbf{Qwen} & \textbf{Mistral}
& \textbf{Qwen} & \textbf{Mistral}
& \textbf{Qwen} & \textbf{Mistral} \\
\midrule
Local
& \textbf{.578} & \textbf{.540}
& 1.60 & \textbf{1.87}
& \textbf{2.27} & \textbf{2.21} \\
Global
& .562 & .530
& \textbf{1.61} & 1.68
& 5.29 & 6.30 \\
PCA
& .574 & .528
& 1.29 & 1.36
& 1160.1 & 1757.4 \\
Random
& .186 & .172
& 1.32 & 1.54
& 134.4 & 160.0 \\
\bottomrule
\end{tabular}
\caption{External evaluation on 210 perturbation pairs pooled across
three subsets. The learned atom frames achieve reconstruction
cosines comparable to PCA, with higher semantic selectivity and
lower nuisance response energies on both models.}
\label{tab:boundary}
\end{table}

Across both models, the learned frames achieve reconstruction
cosines of 0.530--0.578, compared with 0.172--0.186 for Random
and 0.528--0.574 for PCA.
Their advantage over PCA is therefore primarily in selectivity:
Global and Local attain $S/T$ values of 1.60--1.87, exceeding
both baselines, while exhibiting substantially lower nuisance
energies.
Local further reduces nuisance energy relative to Global on both
models, although its $S/T$ improvement is confined to Mistral.
Because absolute nuisance energy depends on frame magnitudes,
we interpret it jointly with $S/T$ and reconstruction cosine.
Together, these results support transfer of semantic selectivity
to external perturbations with minimal surface changes.

\paragraph{Variation across semantic changes.}
The strategy-level analysis shows stronger discrimination for
predicate swaps ($S/T = 2.26$--$4.02$) than for entity swaps
($S/T = 0.65$--$1.15$).
This pattern is consistent with predicate changes altering the
expressed action or relation, whereas entity substitutions can
retain much of the surrounding semantic structure.
The weaker entity-swap results also show that separation is not
uniform across all minimal edits.

This evaluation tests surface-form confusability using perturbations
with assigned SP/SC labels. It does not establish automatic resolution
of genuinely ambiguous meanings. Rather, it shows that the learned
frames retain aggregate semantic selectivity when meaning-preserving
and meaning-changing edits are lexically similar.



\section{Global-to-Local Semantic Geometry}
\label{app:global_local}

Section~3 hypothesizes that semantic geometry is neither fully global nor unconstrainedly local: inputs may modulate the relevance of shared atom directions without redefining the coordinate system itself. We test this by comparing the shared global frame $A_0$ with a locally modulated frame
\[
A(x)=A_0\operatorname{diag}(1+r(x)),
\]
where $r(x)$ is predicted from the anchor representation. This parameterization preserves the directions of the global atoms while adapting their magnitudes to the local semantic neighborhood. Across all seven models, local scaling improves semantic selectivity and reconstruction, whereas stronger input-dependent directional adaptation produces high nominal $S/T$ but collapses semantic reconstruction. The results support a global-to-local organization in which semantic directions are shared while their local importance is context dependent.
We denote the Stage-1 shared frame as $A_0^{(\text{Global})}$ and the Stage-2 
shared frame as $A_0^{(\text{Local})}$ when the distinction matters. Because 
Stage~2 optimizes $A_0$ jointly with the modulation network, 
$A_0^{(\text{Local})}$ may differ from $A_0^{(\text{Global})}$.

\subsection{Full Global and Local Comparison}
\label{app:global_scaling}

Table~\ref{tab:global_scaling_full} reports the complete comparison. Local improves both relative semantic reconstruction error and reconstruction cosine across every evaluated architecture while increasing $S/T$ in all seven cases.

\begin{table}[H]
\centering
\scriptsize
\setlength{\tabcolsep}{3.0pt}
\renewcommand{\arraystretch}{1.06}
\begin{tabular}{llccccc}
\toprule
\textbf{Model} & \textbf{Frame}
& $E_{\mathrm{rec}}\downarrow$
& $\mathrm{Cos}\uparrow$
& $S/T\uparrow$
& $D_{\mathrm{coord}}\downarrow$
& $\mathrm{EffNum}$ \\
\midrule
Mistral
& Global  & 0.701 & 0.703 & 1.27 & \textbf{1.49} & 21.5 \\
& Local & \textbf{0.689} & \textbf{0.714} & \textbf{1.42} & 1.58 & 21.0 \\
\midrule
LLaMA
& Global  & 0.729 & 0.670 & 1.88 & \textbf{1.61} & 23.0 \\
& Local & \textbf{0.719} & \textbf{0.679} & \textbf{1.98} & 1.86 & 23.1 \\
\midrule
Gemma
& Global  & 0.711 & 0.686 & 1.59 & \textbf{2.50} & 21.8 \\
& Local & \textbf{0.696} & \textbf{0.701} & \textbf{2.12} & 2.87 & 21.7 \\
\midrule
Qwen
& Global  & 0.682 & 0.720 & 1.63 & \textbf{1.77} & 22.3 \\
& Local & \textbf{0.667} & \textbf{0.733} & \textbf{1.84} & 1.91 & 22.0 \\
\midrule
GLM
& Global  & 0.716 & 0.683 & 2.05 & \textbf{1.48} & 22.9 \\
& Local & \textbf{0.706} & \textbf{0.692} & \textbf{2.19} & 1.66 & 22.7 \\
\midrule
DeepSeek
& Global  & 0.691 & 0.712 & 1.05 & \textbf{1.84} & 23.0 \\
& Local & \textbf{0.680} & \textbf{0.721} & \textbf{1.26} & 2.05 & 22.3 \\
\midrule
Falcon
& Global  & 0.767 & 0.625 & 1.68 & \textbf{1.45} & 23.6 \\
& Local & \textbf{0.760} & \textbf{0.633} & \textbf{1.79} & 1.67 & 23.5 \\
\bottomrule
\end{tabular}
\caption{Full comparison between the shared global atom frame and local context-dependent diagonal scaling. Local improves semantic reconstruction and $S/T$ across all seven model families while preserving approximately the same effective sparsity.}
\label{tab:global_scaling_full}
\end{table}

The improvement is particularly pronounced for Gemma, where $S/T$ increases from $1.59$ to $2.12$, but is consistent across substantially different architectures, including DeepSeek-MoE ($1.05\!\rightarrow\!1.26$) and Qwen ($1.63\!\rightarrow\!1.84$). The effective number of active atoms remains nearly unchanged, indicating that the gain is not obtained by relaxing sparsity. 
Coordinate drift increases moderately despite improved aggregate selectivity, indicating that the two metrics capture different aspects of SP robustness.
Overall, local adaptation improves semantic selectivity without changing the directional identity of the atoms.

\subsection{Directional Adaptation Does Not Preserve Semantic Coordinates}
\label{app:local_rotation}

To test whether locality should instead modify atom directions, we evaluate two low-rank alternatives,
\[
A_{\mathrm{soft}}(x)=A_0+\gamma U(x)V,
\qquad
A_{\mathrm{QR}}(x)=\operatorname{QR}\!\left(A_0+\gamma U(x)V\right),
\]
where $U(x)\in\mathbb{R}^{d\times r}$ is predicted from the anchor, $V\in\mathbb{R}^{r\times K}$ is learned, $r=8$, and $\gamma=0.05$. LowRank-Soft permits unconstrained directional deformation, whereas LowRank-QR re-orthogonalizes the resulting frame.

The same failure mode appears across all seven architectures. Although both directional variants produce substantially larger nominal $S/T$, relative reconstruction error rises to approximately $0.91$--$1.00$ and cosine similarity falls below $0.35$. In this collapsed reconstruction regime, the large selectivity ratios cannot be interpreted as improved semantic decomposition.

The two variants fail differently. Without re-orthogonalization, the soft low-rank update can increasingly amplify nuisance behavior during optimization. QR prevents this particular loss of orthogonality, but its context-dependent reprojection substantially changes the frame used for projection and reconstruction, and this is associated with severe reconstruction degradation. Diagonal scaling avoids both effects by preserving the shared atom directions and changing only their relative strengths.

These results show that the tested low-rank directional adaptations collapse reconstruction under the present training protocol. This does not establish that semantic locality cannot involve directional changes in general, but indicates that the simpler diagonal parameterization provides a more stable inductive bias for the current setting.

\begin{table}[H]
\centering
\scriptsize
\setlength{\tabcolsep}{3.5pt}
\renewcommand{\arraystretch}{1.05}
\begin{tabular}{llccc}
\toprule
\textbf{Model} & \textbf{Frame}
& $E_{\mathrm{rec}}\downarrow$
& $\mathrm{Cos}\uparrow$
& $S/T\uparrow$ \\
\midrule
Mistral
& Global       & 0.701 & 0.703 & 1.27 \\
& Local      & \textbf{0.689} & \textbf{0.714} & 1.42 \\
& LowRank-QR   & 0.952 & 0.261 & 9.64 \\
& LowRank-Soft & 0.995 & 0.133 & 8.72 \\
\midrule
LLaMA
& Global       & 0.728 & 0.670 & 1.88 \\
& Local      & \textbf{0.719} & \textbf{0.679} & 1.98 \\
& LowRank-QR   & 0.923 & 0.328 & 16.00 \\
& LowRank-Soft & 0.979 & 0.161 & 36.45 \\
\midrule
Gemma
& Global       & 0.711 & 0.686 & 1.59 \\
& Local      & \textbf{0.696} & \textbf{0.701} & 2.12 \\
& LowRank-QR   & 0.910 & 0.344 & 13.58 \\
& LowRank-Soft & 0.973 & 0.180 & 26.45 \\
\midrule
Qwen
& Global       & 0.682 & 0.720 & 1.63 \\
& Local      & \textbf{0.667} & \textbf{0.733} & 1.84 \\
& LowRank-QR   & 0.923 & 0.328 & 14.27 \\
& LowRank-Soft & 0.983 & 0.168 & 19.81 \\
\midrule
GLM
& Global       & 0.716 & 0.683 & 2.05 \\
& Local      & \textbf{0.706} & \textbf{0.692} & 2.19 \\
& LowRank-QR   & 0.919 & 0.331 & 16.27 \\
& LowRank-Soft & 0.973 & 0.174 & 40.57 \\
\midrule
DeepSeek
& Global       & 0.691 & 0.712 & 1.05 \\
& Local      & \textbf{0.680} & \textbf{0.721} & 1.26 \\
& LowRank-QR   & 0.923 & 0.346 & 5.63 \\
& LowRank-Soft & 0.986 & 0.155 & 11.26 \\
\midrule
Falcon
& Global       & 0.767 & 0.625 & 1.68 \\
& Local      & \textbf{0.760} & \textbf{0.633} & 1.79 \\
& LowRank-QR   & 0.926 & 0.325 & 9.81 \\
& LowRank-Soft & 0.973 & 0.169 & 35.36 \\
\bottomrule
\end{tabular}
\caption{Comparison with low-rank directional adaptation. The apparently large $S/T$ values of LowRank-QR and LowRank-Soft coincide with severe reconstruction failure and therefore do not indicate improved semantic geometry.}
\label{tab:rotation_full}
\end{table}

\subsection{Fine-Grained Semantic Changes}
\label{app:local_sc_types}

We next test whether the benefit of local scaling is concentrated in a small subset of semantic changes. Table~\ref{tab:sc_scaling_types} reports Mistral and Gemma as representative examples.

\begin{table}[H]
\centering
\small
\setlength{\tabcolsep}{5pt}
\renewcommand{\arraystretch}{1.05}
\begin{tabular}{lcccc}
\toprule
& \multicolumn{2}{c}{\textbf{Mistral}}
& \multicolumn{2}{c}{\textbf{Gemma}} \\
\cmidrule(lr){2-3}\cmidrule(lr){4-5}
\textbf{SC Type}
& \textbf{Global}
& \textbf{Local}
& \textbf{Global}
& \textbf{Local} \\
\midrule
Entity    & 0.68 & \textbf{0.84} & 0.65 & \textbf{1.14} \\
Attribute & 0.73 & \textbf{0.92} & 0.98 & \textbf{1.64} \\
Relation  & 0.94 & \textbf{1.12} & 1.44 & \textbf{2.10} \\
Negation  & 2.04 & \textbf{2.21} & 2.69 & \textbf{3.35} \\
Quantity  & 1.26 & \textbf{1.41} & 1.60 & \textbf{2.21} \\
Temporal  & 1.65 & \textbf{1.72} & 1.78 & \textbf{2.05} \\
Intent    & 1.03 & \textbf{1.15} & 1.35 & \textbf{1.69} \\
\midrule
All       & 1.27 & \textbf{1.42} & 1.59 & \textbf{2.12} \\
\bottomrule
\end{tabular}
\caption{$S/T$ by semantic-change type for global and locally scaled atom frames. Local improves semantic selectivity broadly rather than specializing to a single SC category.}
\label{tab:sc_scaling_types}
\end{table}

Local improves all seven SC categories for both representative models. The effect is especially clear for subtle substitutions: on Gemma, entity and attribute changes move from $0.65$ and $0.98$ under Global to $1.14$ and $1.64$ under Local. This suggests that local reweighting is particularly useful when the relevant shared semantic directions depend strongly on the surrounding context.

\subsection{Nuisance Suppression Across SP Styles}
\label{app:local_sp_styles}

The improvement in $S/T$ is accompanied by stronger nuisance suppression. Table~\ref{tab:scaling_sp_gemma} reports nuisance projection energy for Gemma across all eight SP styles.

\begin{table}[H]
\centering
\small
\setlength{\tabcolsep}{6pt}
\renewcommand{\arraystretch}{1.05}
\begin{tabular}{lccc}
\toprule
\textbf{SP Style}
& \textbf{Global}
& \textbf{Local}
& \textbf{Reduction} \\
\midrule
Formal     & 8.65  & 2.35 & $3.7\times$ \\
Casual     & 8.79  & 2.44 & $3.6\times$ \\
Question   & 6.44  & 1.75 & $3.7\times$ \\
Technical  & 11.65 & 2.74 & $4.3\times$ \\
Simplified & 9.97  & 2.66 & $3.7\times$ \\
Verbose    & 11.65 & 2.93 & $4.0\times$ \\
Passive    & 12.52 & 3.05 & $4.1\times$ \\
Imperative & 15.42 & 3.11 & $5.0\times$ \\
\bottomrule
\end{tabular}
\caption{Nuisance projection energy for Gemma-2-9B. The Local frame suppresses SP variation consistently across all eight paraphrase styles.}
\label{tab:scaling_sp_gemma}
\end{table}

The reduction ranges from approximately $3.6\times$ to $5.0\times$ and occurs for every style. Local remains effective for verbose, passive, and imperative transformations, which induce some of the largest nuisance displacements under the global frame. The broad reduction argues against the local modulation merely learning a style-specific correction.

\subsection{Training Behavior of Directional Adaptation}
\label{app:local_training}

The reconstruction collapse of the low-rank variants is also visible during optimization. Table~\ref{tab:local_training} reports representative validation statistics from the first and final epochs.

\begin{table}[H]
\centering
\scriptsize
\setlength{\tabcolsep}{3.0pt}
\renewcommand{\arraystretch}{1.05}
\begin{tabular}{llcccc}
\toprule
\textbf{Model} & \textbf{Frame}
& \multicolumn{2}{c}{$L_{\mathrm{val,sem}}$}
& \multicolumn{2}{c}{$L_{\mathrm{val,nui}}$} \\
\cmidrule(lr){3-4}\cmidrule(lr){5-6}
& & \textbf{Epoch 1} & \textbf{Final}
& \textbf{Epoch 1} & \textbf{Final} \\
\midrule
Mistral
& Global       & 0.605 & 0.355 & 1.312 & 0.040 \\
& Local      & 0.380 & 0.342 & 0.031 & 0.010 \\
& LowRank-QR   & 0.715 & 0.616 & 0.716 & 0.332 \\
& LowRank-Soft & 0.767 & 0.616 & 0.014 & 0.257 \\
\midrule
Qwen
& Global       & 0.599 & 0.355 & 1.910 & 0.043 \\
& Local      & 0.385 & 0.338 & 0.034 & 0.017 \\
& LowRank-QR   & 0.753 & 0.635 & 0.825 & 0.376 \\
& LowRank-Soft & 0.816 & 0.636 & 0.108 & 0.195 \\
\midrule
Gemma
& Global       & 0.512 & 0.327 & 1.402 & 0.043 \\
& Local      & 0.357 & 0.310 & 0.033 & 0.011 \\
& LowRank-QR   & 0.622 & 0.518 & 0.721 & 0.250 \\
& LowRank-Soft & 0.681 & 0.520 & 0.024 & 0.151 \\
\midrule
Falcon
& Global       & 0.786 & 0.604 & 1.628 & 0.047 \\
& Local      & 0.637 & 0.590 & 0.032 & 0.013 \\
& LowRank-QR   & 0.961 & 0.863 & 0.828 & 0.351 \\
& LowRank-Soft & 1.023 & 0.869 & 0.029 & 0.217 \\
\bottomrule
\end{tabular}
\caption{Representative training dynamics for global, scaled, and directionally adapted atom frames. LowRank-Soft develops increasing nuisance energy during training, whereas LowRank-QR retains large nuisance error and weak semantic reconstruction.}
\label{tab:local_training}
\end{table}

LowRank-Soft provides the clearest failure mode: nuisance loss increases during training despite the nominal increase in $S/T$. LowRank-QR avoids this growth but retains substantially poorer semantic reconstruction than either Global or Local. These dynamics further indicate that the large selectivity ratios of the directional variants arise in a degraded reconstruction regime rather than from a better decomposition of semantic motion.

\subsection{Reproducibility of Local Modulation}
\label{app:scaling_reproducibility}

Finally, we examine whether context-dependent reweighting is reproducible under optimization stochasticity. We train five runs with different training seeds under a common PCA initialization, align their base atom frames, and then compare both the shared frame and the learned scaling patterns. The scaling vectors $r(x)$ remain strongly correlated across runs, with mean correlations of approximately $0.85$ on Mistral and $0.75$ on Qwen. The base frames are also substantially aligned; for Qwen, the Local runs achieve mean matched atom cosine $0.647$ and subspace similarity $0.820$.

These results indicate that the learned local modulation is not solely an arbitrary consequence of optimization noise. Together with the directional-adaptation results, they support a structured global-to-local picture in which reusable semantic directions are shared globally while local context modulates their relative importance without freely redefining the coordinate system.



\section{Implementation Details}
\label{app:implementation}

\subsection{Representation Extraction and Normalization}
\label{app:implementation1}

For each input $x$, we extract the raw final-token hidden
state $h_\ell(x)\in\mathbb{R}^d$. The representation map
used in the main paper includes per-dimension
standardization:
\[
[\Phi_\ell(x)]_j
=
\frac{[h_\ell(x)]_j-\mu_j}{\sigma_j},
\qquad j=1,\ldots,d,
\]
where $\mu_j$ and $\sigma_j$ are per-dimension statistics estimated over all
extracted representations of the corpus (anchors and all SP/SC variants) at
the target layer. These statistics do not use the SP/SC labels and are
shared by all methods and baselines. Inputs are tokenized with left padding
and truncated to at most $64$ tokens, and $h_\ell(x)$ is the hidden state of
the final token.
Representations are
extracted in half precision and subsequently processed
in floating point for atom learning. Causal interventions
convert reconstructed displacements back to raw
hidden-state units as described in Appendix~\ref{app:causal}.

\begin{table}[H]
\centering
\caption{Primary representation layer used for each model family.}
\label{tab:primary_layers}
\begin{tabular}{lc}
\toprule
\textbf{Model} & \textbf{Layer} \\
\midrule
Mistral-7B-Instruct-v0.3  & 20 \\
LLaMA-3.1-8B-Instruct     & 16 \\
Gemma-2-9B                 & 12 \\
Qwen2.5-7B-Instruct       & 20 \\
GLM-4-9B-Chat              & 20 \\
DeepSeek-MoE-16B-Chat      & 20 \\
Falcon3-7B                 & 12 \\
\bottomrule
\end{tabular}
\end{table}

Unless otherwise specified, we use $K=256$ atoms and a sparsity budget of $k=32$. Target layers are selected independently for each architecture based on the layer-wise analysis reported in Appendix~B. The same selected layer is used for the Global and Local variants and their corresponding baselines.

\subsection{Atom Frame Training}
\label{app:atom_training}

The Global model learns a shared atom frame $A_0\in\mathbb{R}^{d\times K}$ together with a linear sparse encoder
\[
\tilde{\alpha}=W\Delta z+b,
\qquad
\alpha=\operatorname{TopK}(\tilde{\alpha},k),
\]
where TopK retains the $k$ coefficients with largest absolute magnitude while preserving their signs. 
During backpropagation, gradients propagate through the retained coefficient values, while the discrete TopK index selection is treated as non-differentiable; consequently, coefficients outside the selected support receive zero gradient from the reconstruction path for that example. 
The atom frame is initialized with the leading \(K\) principal directions of the training SC displacements, with $W=A_0^\top$ and $b=0$; $A_0$ and $W$ are subsequently optimized independently.

The implementation uses element-wise MSE reductions.
In terms of the geometric energies defined in Section 3.3,
the gradient-contributing objective is
\[
\mathcal{L}_{\mathrm{impl}}
=
\frac{1}{d}\mathcal{L}_{\mathrm{sem}}
+
\frac{0.5}{K}\mathcal{L}_{\mathrm{nui}}
+
\frac{0.3}{K}\mathcal{L}_{\mathrm{stress}}
+
\frac{0.3}{K^2}\mathcal{L}_{\mathrm{orth}}.
\]
Thus, the numerical weights $1.0$, $0.5$, $0.3$, and
$0.3$ apply to the corresponding element-wise
mean-squared errors.

The implementation additionally includes the
activation-frequency statistic
\[
\bar u_j
=
\frac{1}{N}\sum_{n=1}^{N}
\mathbf{1}[|\alpha_{n,j}|>\epsilon],
\qquad
\mathcal{L}_{\mathrm{bal}}
=
\operatorname{Var}(\bar u_1,\ldots,\bar u_K),
\]
with coefficient $0.1$ in the reported total.
The hard indicator supplies no parameter gradient,
so this statistic monitors utilization without
providing a gradient-based balancing penalty.

We train for 100 epochs using AdamW with learning rate $5\times10^{-4}$ and weight decay $10^{-5}$. A cosine-annealing schedule reduces the learning rate to $10^{-6}$, gradients are clipped at norm $1.0$, and validation is performed every 10 epochs. Unless otherwise stated, experiments use seed 42. The standard split uses even-indexed SP/SC transformations for training and odd-indexed transformations for evaluation; stricter neighborhood-disjoint and perturbation-generator transfer protocols are described separately in Appendix~\ref{app:generalization}. 
Every 10 epochs, the current model is evaluated on the odd-indexed held-out split, and the checkpoint with the best validation criterion is retained for reporting. Thus, the held-out split is not used for gradient updates, but it is used for checkpoint selection as well as final evaluation.

\subsection{Local Modulation Model}

The Local model is initialized from the trained Global solution and introduces anchor-dependent diagonal modulation,
\[
A(x)=A_0\operatorname{diag}(1+r(x)),
\qquad
r(x)=\tanh(g_\eta(z)).
\]
The modulation network $g_\eta$ is an MLP with architecture
$d\rightarrow512\rightarrow512\rightarrow K$, using LayerNorm and GELU activations. Its output layer is initialized to zero, so that $r(x)=0$ and $A(x)=A_0$ at the beginning of local training. This initialization makes the Local model a continuation of the learned Global geometry rather than an independently initialized coordinate system.

During local training, $A_0$, the sparse encoder $(W,b)$, and $g_\eta$ are optimized jointly using the same objective and optimization settings as the Global model. 
Because the adaptation is diagonal, local conditioning changes atom magnitudes but does not introduce anchor-specific rotations. We refer to the two model variants as Global and Local throughout the paper. 

\subsection{Baselines and Sparse Encoding}

We compare the learned atom representation with PCA, Random, and reconstruction-only baselines. For PCA, $A_0$ is fixed to the leading $K=256$ principal directions estimated from training semantic-changing displacements. For Random, $A_0$ is a $K$-dimensional random orthonormal frame obtained by QR factorization of a Gaussian random matrix. In both cases, the atom frame is frozen while the sparse encoder is trained under the corresponding experimental protocol.

The Recon-Only baseline uses the same learnable atom frame and TopK encoder as the Global model but optimizes only $\mathcal{L}_{\mathrm{sem}}$. It therefore provides an SAE-style sparse reconstruction control without semantic-preserving supervision. Objective ablations additionally remove individual invariance and structural terms while leaving the architecture unchanged; complete results are reported in Appendix~B.

For sparsified signatures, TopK is applied to the dense $K$-dimensional atom coordinates and all non-selected entries are set to zero. No additional $\ell_1$ sparsity penalty is used, since the TopK operator directly enforces $\|\alpha\|_0\leq k$. Sparsity sweeps retrain the model independently for each
$k\in\{4,8,16,24,32,48,64\}$.

\subsection{Fine-Tuned and Knowledge-Distilled Model Variants}
\label{app:model_variants}

To test whether the learned semantic coordinate system persists under model modification, we construct two adapted variants of each evaluated base model: a supervised LoRA fine-tuned variant and a knowledge-distilled LoRA variant. These checkpoints are used only in the model-modification experiments; the atom frames and downstream semantic classifiers are learned from the original base models and are not re-estimated on the adapted variants.

Each base model is adapted on a 10K-example subset of the Alpaca instruction-following dataset. Fine-tuning uses LoRA with rank $r=8$, scaling parameter $\alpha=32$, dropout $0.05$, and, where supported by the architecture, adapters on the query and value projection layers. The base model weights remain frozen. Training uses 4-bit NF4 quantization, learning rate $2\times10^{-4}$, micro-batch size 2 with 16 gradient-accumulation steps, maximum sequence length 256, and 300 optimization steps. The supervised variant is optimized using token-level cross-entropy.

The knowledge-distilled variant uses the same LoRA parameterization and training data, but augments the supervised objective with teacher matching. For each architecture, the unmodified base checkpoint serves as a frozen teacher loaded in 8-bit precision, while a 4-bit copy equipped with trainable LoRA adapters serves as the student. The student is optimized using
\[
\mathcal{L}_{\mathrm{KD}}
=
0.5\,\mathcal{L}_{\mathrm{CE}}
+
0.5\,\mathcal{L}_{\mathrm{KL}},
\]
where $\mathcal{L}_{\mathrm{KL}}$ matches the student and teacher output distributions with temperature $T=2$. The distilled variants use the same 300-step training schedule as the supervised fine-tuned variants. After training, the teacher is discarded and representations are extracted from the adapted student checkpoint. Unlike compression-oriented distillation, the teacher and student retain the same underlying architecture; this condition therefore tests robustness to teacher-regularized parameter adaptation rather than architectural compression.

For evaluation, the Global atom frame learned from the base checkpoint is applied unchanged to representations extracted from the fine-tuned and knowledge-distilled variants. The SC-type classifier is likewise trained only on base-model signatures and evaluated zero-shot on the corresponding adapted-model signatures. Preservation of classification accuracy therefore measures whether the semantic coordinate organization learned in the base model remains aligned after parameter adaptation.

We additionally measure pairwise signature drift between corresponding base and adapted examples using cosine similarity and Euclidean distance. For model-source classification, equal-sized signature pools from the base, fine-tuned, and knowledge-distilled variants are randomly divided into 80/20 training and validation splits. These evaluations probe complementary properties: zero-shot SC classification measures preservation of semantic organization, whereas signature drift and source classification quantify residual sensitivity to the underlying model modification.

\subsection{Downstream Classifiers and Retrieval Evaluation}

For semantic-change classification, atom signatures are used to predict the seven SC transformation types: entity, attribute, relation, negation, quantity, temporal, and intent. The classifier is a two-hidden-layer MLP with architecture
$\mathrm{dim}_{\mathrm{in}}\rightarrow128\rightarrow128\rightarrow7$, ReLU activations, and dropout $0.2$. It is trained for 100 epochs on base-model signatures using the training split and evaluated on held-out signatures. For model-modification experiments, this classifier is trained only on the base checkpoint and applied without retraining to fine-tuned and knowledge-distilled variants.

For model-source analysis, the same MLP architecture is used with a three-class output corresponding to base, fine-tuned, and knowledge-distilled variants. Equal-sized signature pools from the three conditions are randomly divided into 80/20 training and validation partitions. Signature stability is additionally measured directly using cosine similarity and Euclidean distance between corresponding examples before and after model modification.

For semantic retrieval, each SP representation is used as a query and anchor representations are ranked by cosine similarity. A retrieval is correct when the matched anchor belongs to the same semantic neighborhood as the query. We report Recall@1, Recall@5, mean average precision, and AUROC. All compared representations use the same query--anchor pairs so that differences reflect the representation rather than the retrieval protocol.

\subsection{Compute Resources}

All experiments were conducted on NVIDIA A100 40GB GPUs. Representation 
extraction (forward pass through the frozen backbone for all 64{,}093 
sentences at each target layer) takes approximately 20 to 30 minutes per 
model. Atom frame training (100 epochs, Global or Local) takes 
approximately 2 hours per model at the primary layer. The sparsity sweep 
($k \in \{4, 8, 16, 24, 32, 48, 64\}$) and layer-wise analysis (10 layers) 
each require one training run per setting. Across all seven model families 
and all reported experiments, including atom-frame and baseline comparisons, sparsity 
sweeps, layer-wise analyses, local conditioning variants, cross-seed 
reproducibility (5 seeds), held-out generalization splits, causal 
interventions, and downstream evaluations, the total compute is approximately 
350 GPU hours. The language-model backbones remain frozen throughout atom learning; atom 
learning optimizes only the $K$-atom frame, sparse encoder, and (for the 
Local model) the lightweight modulation network.
